\documentclass[sigconf]{acmart}

\usepackage{multirow}
\usepackage{subfigure}

\usepackage{enumitem}
\usepackage{pifont}
\usepackage{balance}

\usepackage{algpseudocode}
\usepackage{algorithm}

\usepackage{threeparttable}
\usepackage{enumitem}

\usepackage{booktabs}
\usepackage[normalem]{ulem}
\useunder{\uline}{\ul}{}

\AtBeginDocument{%
  }

\copyrightyear{2025}
\acmYear{2025}
\setcopyright{acmlicensed}
\acmConference[KDD '25] {Proceedings of the 31st ACM SIGKDD Conference on Knowledge Discovery and Data Mining V.2}{August 3--7, 2025}{Toronto, ON, Canada.}
\acmBooktitle{Proceedings of the 31st ACM SIGKDD Conference on Knowledge Discovery and Data Mining V.2 (KDD '25), August 3--7, 2025, Toronto, ON, Canada}
\acmISBN{979-8-4007-1454-2/25/08}
\acmDOI{10.1145/3711896.3737047}

\begin{document}

\title{MetaEformer: Unveiling and Leveraging Meta-patterns for Complex and Dynamic Systems Load Forecasting}

\author{Shaoyuan Huang}
\authornote{Both authors contributed equally to this research.}
\orcid{0000-0002-4091-6457}
\affiliation{%
  \institution{Tianjin University}
  \city{Jinnan Qu}
  \state{Tianjin Shi}
  \country{China}}
\email{hsy_23@tju.edu.cn}

\author{Tiancheng Zhang}
\authornotemark[1]
\orcid{0009-0006-7681-1223}
\affiliation{%
  \institution{Tianjin University}
  \city{Jinnan Qu}
  \state{Tianjin Shi}
  \country{China}}
\email{tianchengzhang@tju.edu.cn}

\author{Zhongtian Zhang}
\orcid{0009-0002-6724-2226}
\affiliation{%
  \institution{Tianjin University}
  \city{Jinnan Qu}
  \state{Tianjin Shi}
  \country{China}}
\email{zzt_2022@tju.edu.cn}

\author{Xiaofei Wang}
\authornote{Corresponding author.}
\orcid{0000-0002-7223-1030}
\affiliation{%
  \institution{Tianjin University}
  \city{Jinnan Qu}
  \state{Tianjin Shi}
  \country{China}}
\email{xiaofeiwang@tju.edu.cn}

\author{Lanjun Wang}
\orcid{0000-0002-7696-5330}
\affiliation{%
  \institution{Tianjin University}
  \city{Jinnan Qu}
  \state{Tianjin Shi}
  \country{China}}
\email{wang.lanjun@outlook.com}

\author{Xin Wang}
\orcid{0000-0001-9651-0651}
\affiliation{%
  \institution{Tianjin University}
  \city{Jinnan Qu}
  \state{Tianjin Shi}
  \country{China}}
\email{wangx@tju.edu.cn}

\renewcommand{\shortauthors}{Shaoyuan Huang et al.}

\begin{abstract}
Time series forecasting is a critical and practical problem in many real-world applications, especially for industrial scenarios, where load forecasting underpins the intelligent operation of modern systems like clouds, power grids and traffic networks.
However, the inherent complexity and dynamics of these systems present significant challenges. Despite advances in methods such as pattern recognition and anti-non-stationarity have led to performance gains, current methods fail to consistently ensure effectiveness across various system scenarios due to the intertwined issues of complex patterns, concept-drift, and few-shot problems.
To address these challenges simultaneously, we introduce a novel scheme centered on fundamental waveform, a.k.a., meta-pattern. Specifically, we develop a unique Meta-pattern Pooling mechanism to purify and maintain meta-patterns, capturing the nuanced nature of system loads. Complementing this, the proposed Echo mechanism adaptively leverages the meta-patterns, enabling a flexible and precise pattern reconstruction. Our Meta-pattern Echo transformer (MetaEformer) seamlessly incorporates these mechanisms with the transformer-based predictor, offering end-to-end efficiency and interpretability of core processes. Demonstrating superior performance across eight benchmarks under three system scenarios, MetaEformer marks a significant advantage in accuracy, with a 37\% relative improvement on fifteen state-of-the-art baselines. 
\end{abstract}





\maketitle

\newcommand\kddavailabilityurl{https://doi.org/10.5281/zenodo.15502862}

\ifdefempty{\kddavailabilityurl}{}{
\begingroup\small\noindent\raggedright\textbf{KDD Availability Link:}\\
% please change the following context to include multiple artifacts if necessary.
The source code of this paper has been made publicly available at \url{\kddavailabilityurl}.
\endgroup
}

\section{Introduction}

Time series forecasting, a fundamental task in machine learning, has long been the subject of extensive research across a multitude of domains such as the industry failure maintenance~\cite{DALZOCHIO2020103298}, medical diagnosis~\cite{Hu2022ATD}, and weather nowcasting~\cite{zhang2023skilful}. This focus stems from its ability to predict relevant figures in advance, assisting in decision-making and formulating response strategies. As technology advances, an increasing number of modern systems (like cloud computing platforms \cite{10.1145/3580305.3599453, 10.1145/3543507.3583436, 9969934}, power grids \cite{Yang2023, gasparin2022deep}, and traffic networks \cite{10.1145/3580305.3599528, 10.1145/3534678.3539250}) have integrated load forecasting capabilities into their operational processes. This integration marks a shift of systems from mere functionality to intelligent services.

\begin{figure}[t]
    \centering
    \setlength{\belowcaptionskip}{-0.5cm}
    \includegraphics[width=\columnwidth]{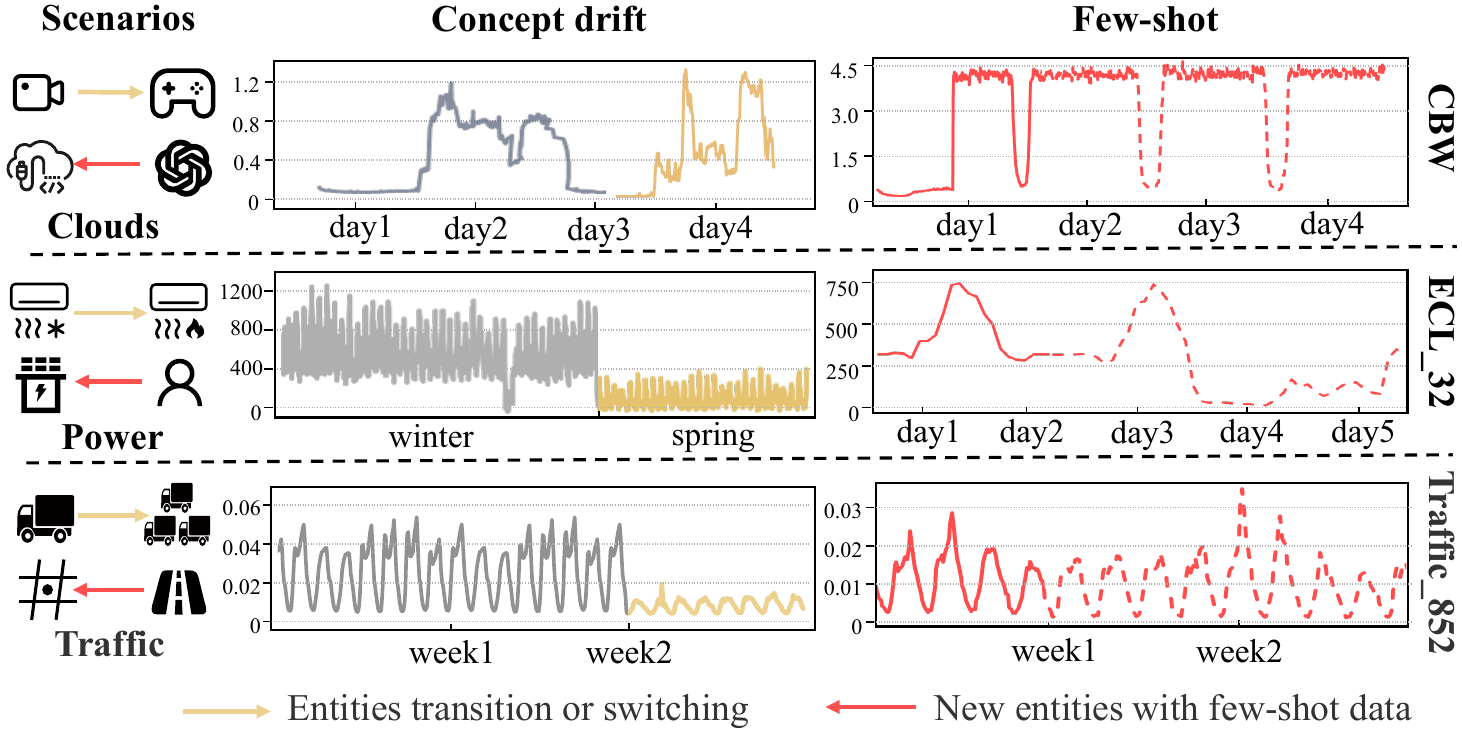}   
    \caption{Systems loads variations due to entity behaviors.}
    \label{fig:scence_analysis}
    \Description{Systems loads variations due to entity behaviors.}
\end{figure}

However, these systems, which are characterized by \textbf{complexity} and \textbf{dynamics}, introduce significant challenges for load forecasting \cite{10.1145/3534678.3542612}. Specifically, the system \textbf{complexity} arises from its diverse internal entities and external influences like natural, economic, and social factors. For instance, the multitude of servers and applications in cloud platforms; similarly, in traffic systems, it manifests through the varying characteristics of different roads and traffic patterns. The confluence of complexity exhibits intricate patterns with seasonality, trend, outlier, and noise, which tests the latent pattern recognition capabilities of predictors~\cite{10.1145/3534678.3542612}.

% 另一方面，现实世界的系统通常具有高度的动态性，这是由于系统中的实体随着系统的运行和与外界的交互，状态模式逐渐发生变化，甚至实体数量也会不断更新。例如在云服务系统中不断发生的应用调度与部署，电力系统中负载随季节变化的过程。对于负载预测来说，系统的动态性往往会比复杂性更具挑战，因为其会导致负载分布发生漂移或引入完全未知的分布，这对模型对抗这种非平稳性数据（鲁棒性）和适应未知模式（泛化性）的能力提出了较高的要求。

% 具体来说，导致动态性的实体行为可以被总结为以下两种：1.实体的迁移的和切换，如图1(a)所示，我们可以观察到在云服务系统中由于频繁的应用切换会导致设备的负载出现模式变化，类似的现象也出现在电力系统中的季节迁移和路网系统中的traffic congestion transition等情况中。在时序预测任务中，数据分布随时间变化的现象被识别为concept-drift\cite{8496795}，这时的模型往往会因为过度关注旧有模式而未能及时捕获切换后的数据模式而导致效果下降。

On the other hand, real-world systems exhibit extremely high dynamics, a characteristic stemming from the evolving state of systems entities as they operate and interact with the external environment, which can be attributed to two types of entity behaviors:

\textbf{(i) Entity transition and switching.} As illustrated in Fig. \ref{fig:scence_analysis}, frequent application switches in cloud platforms could cause load pattern changes, similar to seasonal migration in power grids and traffic congestion transitions in road networks. In time series forecasting, the phenomenon of time-variant data distribution is identified as \textbf{concept-drift} \cite{8496795}. In this context, models often experience performance degradation due to their focus on outdated patterns, failing to promptly capture the post-switch patterns \cite{kim2021reversible, fan2023dish, Chen2024}.

% 2. 新实体的加入。如图1(b)所示，系统中新增的实体可能会引入未知的负载模式，例如云系统中新增的应用类型，电力系统中新接入的工厂用户以及路网系统中新增的高速路段。在这种情况下，模型需要从少量的(few-shot)新增实体数据中捕获新的负载模式，这对模型依据已有模式重构新模式的泛化能力提出了考验。
\textbf{(ii) Introduction of new entities.} As Fig. \ref{fig:scence_analysis} shows, newly added entities may bring unknown load patterns. This could be new types of application in cloud platforms, newly connected industrial users in power grids, or new segments in traffic networks. In such scenarios, models are challenged to capture new patterns from the limited data of the \textbf{few-shot} \cite{brinkmeyer2022few}, new entities, testing their ability to generalize and reconstruct new patterns from existing knowledge.

\begin{table*}[htbp]
  \centering
  \caption{Comparison of different types of methods, focusing on advantages applicable to complex and dynamic series forecasting.}
    \begin{tabular}{c|cccc}
    \hline
    Methods & End-to-End & Pattern Recognition & Dynamics Adaptation & Interpretability \\
    \hline
     Generic Models~\cite{nie2022time, yi2024filternet, ekambaram2023tsmixer,  yi2024frequency, zhang2023crossformer} & \ding{51}      & \ding{51}     & \ding{55}     & \ding{55} \\
    Decomposition-based~\cite{zeng2023transformers, wu2021autoformer, zhou2022fedformer, 10.1145/3534678.3539234} & \ding{51}     & \ding{51}      & \ding{55}     & \ding{55} \\
    Anti-non-stationarity~\cite{liu2022non, liu2023adaptive, liu2023koopa} & \ding{51}      & \ding{55}     & \ding{51}      & \ding{55}  \\
    Clustering-based~\cite{8457781,jayakumar2020self, Yang2023, bitencourt2023embedding, 10.1145/3580305.3599453} & \ding{55}     & \ding{51}      & \ding{51}      & \ding{51}  \\
    Ours  & \ding{51}      & \ding{51}      & \ding{51}      & \ding{51}  \\
    \hline
    \end{tabular}%
  \label{tab:dis_compapre}%
\end{table*}%

Recent research in time series forecasting has witnessed significant strides, with some methods addressing certain challenges in system load forecasting: generic deep learning models excel at pattern recognition, capturing complex load patterns, while \emph{decomposition-based methods} \cite{wu2021autoformer, 10.1145/3534678.3539234, zhou2022fedformer} further enhance this by directly utilizing latent pattern information. \emph{Anti-non-stationary approaches} \cite{liu2022non, liu2023adaptive, liu2023koopa} demonstrate robustness to load variability, actively adapting to dynamic factors like frequent entity transitions. \emph{Clustering-based strategies} \cite{Yang2023, jayakumar2020self, bitencourt2023embedding}, commonly used in industrial settings, leverage pre-constructed model pools to extract targeted load patterns and adapt dynamically through model switching.

% 尽管有了如此多的努力，这些工作each manifest inherent limitations in certain respects，仍然没有一个工作可以同时tackle the challenges of complexity, concept drift, and few-shot simultaneously。另外，模型的可解释性也是真实工业场景下不容忽视的需求（尤其是在动态系统的context下）。

Despite these efforts, existing methods each exhibits inherent limitations when attempting to tackle the challenges of complexity, concept drift, and few-shot scenarios simultaneously. Furthermore, the lack of interpretability remains a key concern in real-world industrial contexts, especially within dynamic systems.

This paper's goal is to concurrently address the complexity and dynamics challenges of systems load forecasting, while providing the efficiency and interpretability required for real-world systems. Inspired by Wavelet Transform principles \cite{sifuzzaman2009application}, we suggest that all load patterns and their variations in a system are composed of fundamental waveforms. To achieve this, such waveform should exhibit fine granularity and be highly representative, going beyond traditional periodic patterns to form a 'pattern within patterns', a.k.a., meta-pattern. Based on this idea, we design the \textbf{\emph{Meta-pattern Pooling}} and a novel \textbf{\emph{Echo mechanism}} to enhance the transformer-based predictors' focus on the essential differences and connections in complex and dynamic load patterns. % performance in various system scenarios.

% 为达到这一点，这种基础波形需要有着超越传统周期模式的精细粒度和高代表性，使其成为一种模式中的模式，即元模式。

% need to have a fine granularity and high representativeness beyond the usual cyclic patterns, making it a pattern within a pattern, i.e., a meta-pattern.

The Meta-pattern Pooling design draws inspiration from wavelet basis, which are fundamental elements that exhibit temporal locality and form all vectors in a space. The pooling mechanism integrates time series decomposition and threshold-based pattern purification to isolate these basic units (meta-patterns) from load series and pool them into the meta-pattern pool (MPP) for maintenance.

% The pooling mechanism purifies and maintains these basic units (meta-patterns), integrating time series decomposition and alignment-based pattern recognition to isolate core components from load series and aggregate them into the meta-pattern pool (MPP).

% Therefore, we integrate time series decomposition and aligned-based pattern recognition to isolate fine-grained core components from load series and aggregate them into the meta-pattern pool (MPP).

% The design of Meta-pattern Pooling is inspired by the wavelet basis, which are fundamental elements capable of forming all vectors in a space and exhibit temporal locality. In our design, the primary goal of the Meta-pattern Pooling is to purify and maintain these basic units (meta-patterns) to guide subsequent forecasting. Therefore, we integrate time series decomposition and aligned-based pattern recognition to isolate fine-grained core components from load series and aggregate them into the meta-pattern pool (MPP).

The Echo mechanism is designed to deconstruct the target load series and reconstruct load patterns, the concept is akin to the process in wavelet transform where the wavelet basis is used to analyze the original signal. Specifically, the Echo enhances complex pattern recognition by integrating relevant meta-patterns. For concept-drift problems, it identifies post-transition patterns from the MPP, mitigating the model's reliance on outdated information. For few-shot issues, the Echo combines existing meta-patterns in the pool to reconstruct unknown patterns, enabling rapid adaptation to unknown data distributions.

% Both the Meta-pattern Pooling and the Echo mechanism are designed as learnable layers with adaptive parameters, which means they can be directly plugged into encoder-decoder architecture as evolving companion modules. This approach avoids pre-built structures, prevents information leakage and maintains long-term effectiveness. Our framework, namely Meta-pattern Echo transformer (MetaEformer), excels in various system load forecasting tasks, delivering a seamless end-to-end process and scenario-focused interpretability to meet industrial requirements.

Both mechanisms are implemented as learnable layers with adaptive parameters, allowing them to be directly integrated into the encoder-decoder architecture as evolving companion modules. This avoids pre-built structures, prevents information leakage, and ensures long-term effectiveness. Our framework, Meta-pattern Echo transformer (\textbf{MetaEformer}), excels in various system load forecasting tasks, delivering seamless end-to-end processing and interpretability to meet industrial demands.

To summarize, our main contributions are as follows:
\begin{itemize}[leftmargin=*]
\item We introduce a new perspective on meta-patterns and propose
a novel \emph{Meta-pattern Pooling} mechanism for purifying and maintaining these fundamental waveforms in system loads. This mechanism can autonomously evolve alongside the model training and inference, capturing the intricacies of load series.
% This approach enhances the understanding and representation of load series.
\item To leverage the meta-patterns, we propose an innovative \emph{Echo} mechanism that deconstructs load series and reconstructs patterns. This mechanism simultaneously adapts to load forecasting challenges like complex pattern, concept-drift, and few-shot problems, offering a more flexible and precise 
forecasting model.
\item Our framework, MetaEformer, integrates these mechanisms as learnable and adaptive components into the transformer-based predictor. This design not only achieves superior performance on eight real-world system load datasets, but also offers end-to-end efficiency and clear interpretability of core processes.
\end{itemize}

\section{Related Work}\label{sec:rw}
In this section, we review previous research regarding the aforementioned decomposition-based, anti-non-stationary and clustering-based approaches.

\textbf{Decomposition-based methods} integrate time series decomposition with advanced predictors to explicitly identify patterns by extracting the key components (such as the seasonal component) from the original time series. Wu et al. \cite{wu2021autoformer} introduces traditional decomposition method \cite{cleveland1990stl} and auto-correlation into the transformer for superior complex long-term forecasting. Following this paradigm, other works also integrate various decomposition techniques. Examples include the frequency-based FEDformer \cite{zhou2022fedformer} and DLinear \cite{zeng2023transformers}, and the period and phase-focused Quatformer \cite{10.1145/3534678.3539234}.

% the 2D convolutional decomposition-based Timesnet \cite{wu2022timesnet}.

These methods are evaluated as effective in extracting predictive information from complex time series (especially for system loads in their experiments). However, from a dynamism perspective, their effectiveness diminishes, especially when facing changing or unknown load patterns, since over-reliance on historical patterns could misleadingly direct the model.

% 遵循这一范式，越来越多的工作开始引入其他分解技术并依此提出更有效的预测框架，例如基于频率分解的fedformer，基于周期与相位的Quatformer，基于2D卷积分解的Timesnet.

% 这些方法均被验证可以有效地从复杂的时间序列（尤其是真实系统负载）中提取更有助于预测的信息，但从系统的动态性角度来看，此类方法变得没有那么有效。尤其是在面临实体切换导致的负载模式变化或新实体导致的未知模式时，过度依赖局部的历史模式反而会将模型引入错误的方向。

% 缺少与系统负载预测的联系
\textbf{Anti-non-stationary approaches} are emerging techniques that focus on analyzing loads' time-variant characteristics and developing methods to eliminate and reconstruct non-stationary aspects, enhancing models' resilience in dynamic scenarios. 
% Liu et al. \cite{liu2022non} initially points out that traditional preprocessing such as first-order differencing, might weaken the attention mechanism in models like Transformers when dealing with non-stationary data. Therefore, they propose a "stabilize then destabilize" approach with a destabilizing attention mechanism, enhancing the model's adaptation to non-stationary data. Similarly, Liu et al. \cite{liu2023adaptive} also highlights that stabilization preprocessing may overlooks the different statistical characteristics among time series segments. In response, they propose a segment-based method with a lightweight network to adaptively address non-stationarity across different segments. 
% Liu et al. \cite{liu2022non} highlight how traditional preprocessing, like first-order differencing, could impair Transformers' effectiveness on non-stationary data, suggesting a "stabilize then destabilize" method with a destabilizing attention mechanism for better adaptation. Similarly, Liu et al. \cite{liu2023adaptive} note the potential oversight of varying statistical features in time series segments by stabilization techniques, advocating for a segment-based approach with a lightweight network to flexibly handle non-stationarity across segments.

Liu et al. \cite{liu2022non} propose a "stabilize then destabilize" method, addressing the limitations of traditional preprocessing like first-order differencing in Transformers, using a destabilizing attention mechanism for better adaptation to non-stationary data. Similarly, Liu et al. \cite{liu2023adaptive} note the potential oversight of varying statistical features by stabilization techniques, advocating for a segment-based approach with a lightweight network to handle non-stationarity series.

While these methods refine traditional stabilization preprocessing methods, they still deconstruct the inputs directly and risk losing key non-stationarity details crucial for real-world systems. Besides, these methods do not provide a clear interpretability of how the system dynamics are represented and perceived.

\textbf{Clustering-based strategies} enhance model robustness by maintaining multiple data and model pools that jointly contribute to forecasting. This kind of paradigm is commonly employed for system-oriented forecasting tasks. In the cloud platforms, both Kim et al. \cite{8457781} and Jayakumar et al. \cite{jayakumar2020self} introduce multiple model-based predictions using clusters of computational loads. Similarly, in the power load domain, concepts akin to data pooling have been implemented in work \cite{Yang2023}.

% The clustering-based strategies offer two main benefits. Firstly, the multiple data clusters enable statistical pattern recognition. Different from decomposition-based methods, these approaches explicitly reveal the types of data distributions, aiding users in better understanding system's load patterns. Secondly, regarding system dynamics, models switching within the framework can better interpret dynamic behaviors like entity transitions and the introduction of new entities \cite{}.
The clustering-based strategies yield two primary advantages. First, they facilitate statistical pattern recognition through data clusters, thereby enhancing comprehension of system load patterns. Second, they adeptly interpret dynamic behaviors, such as entity transitions and the integration of new entities, through model switching within the framework \cite{bitencourt2023embedding}.
%distinct from decomposition-based methods, by clearly exposing data distribution types, 
However, the primary drawback of these methods is their complexity. Discarding the end-to-end pipeline for multiple data clusters and models escalates computational expenses. Despite Huang et al. \cite {10.1145/3580305.3599453} making efforts to incorporate the capabilities of multiple models into a single framework, it is still not completely end-to-end. Their reliance on full historical data for global pools raises concerns about potential information leakage.

% 尽管Huang et al \cite{10.1145/3580305.3599453}在将多个模型的能力融入到一个Transofmer模型中做出了巨大的努力，但其仍没有做到彻底的端到端，and their reliance on full historical data for data pools constructing raises concerns about potential information leakage.

% Additionally, their reliance on full historical data for data pools constructing raises concerns about potential information leakage.

% 如表1所示，通过引入meta-patterns这一创新概念和相应的机制，MetaEformer以全新的计算方式成功吸纳（实现，再现并融合的意思）了各类方法的核心优势，避免了其固有弊端，从而使它最适配于系统负载预测任务。
\textbf{In summary}, as shown in Table \ref{tab:dis_compapre}, leveraging the innovate meta-patterns, MetaEformer harnesses the core strengths of various approaches while overcoming their inherent limitations, making it particularly well-suited for system load forecasting tasks.

\section{Notations and Problem Definition}\label{npd}

\textbf{System load forecasting.} A load series comprises a sequence of load points (e.g., cloud server bandwidth, electricity consumption or traffic flow) generated over time $t$, denoted as follows:
\begin{equation}
   \quad \mathcal{X} = \{x_{t_1}, x_{t_2}, \ldots, x_{t_n}\ |\ x_{t} \in \mathbb{R}^d \} 
\end{equation}where $d$ is the feature dimension with $d\geq 1$. These data points typically appear at fixed time intervals $k$, and given a look-back window of size $L$, the objective is to predict the data for the next $L_y$ intervals. Therefore, for a forecasting task starting from $t$, the input can be represented as $\mathcal{X}_t = \left(x_{i}\right)_{i=t-L*k}^{t}$, and the target is to predict the future segment $\mathcal{Y}_t = \left(x_{i}\right)_{i=t+k}^{t+L_y*k}$.

\textbf{Concept-drift and Few-shot.} Traditional time series forecasting assumes that $x_i$ in $\mathcal{X}_t$ and $\mathcal{Y}_t$ follow the same distribution $\mathcal{D}_t$, i.e., $x_i \sim \mathcal{D}_t$. Besides, historical data distributions $D$ are presumed to encompass any potential future distributions. However, under concept-drift, $x_i$ may follow multiple distributions, for instance, $\left(x_{i}\right)_{i=t-L*k}^{S1} \sim \mathcal{D}^1_t$ and $\left(x_{i}\right)_{i=S1}^{S2} \sim \mathcal{D}^2_t$, and $S1$ is the time for entity transition. On the other hand, in few-shot scenarios, a future distribution $\mathcal{D}_t^{'}$ may not belong to any distribution learned by the model, signified as $\mathcal{D}_t^{'} \notin D$. Both scenarios significantly increase the difficulty of forecasting.

\section{Methodology}\label{section:MetaEformer}
% In this section, we will introduce the the overall architecture of MetaEformer, as shown in Fig. \ref{fig:framework architecture}, including the Meta-pattern Pooling mechanism and the forecasting framework. 

% 在这一部分我们将着重介绍xx和MetaEformer的整体架构。正如前文所说，XX被设计成可以集成在端到端模型中的一个组件来辅助模型应对三个行为所导致的复杂性与非平稳性。

In this section, we introduce the overall architecture of MetaEformer, and dive into the proposed Meta-pattern Pooling and Echo mechanism. The two mechanisms are designed as inherently interpretable components and integrated into the encoder-decoder architecture, ensuring an end-to-end pipeline while handling the complexity and dynamics caused by systems entities.

\begin{figure*}[ht]
    \centering
    \includegraphics[width=0.78\textwidth]{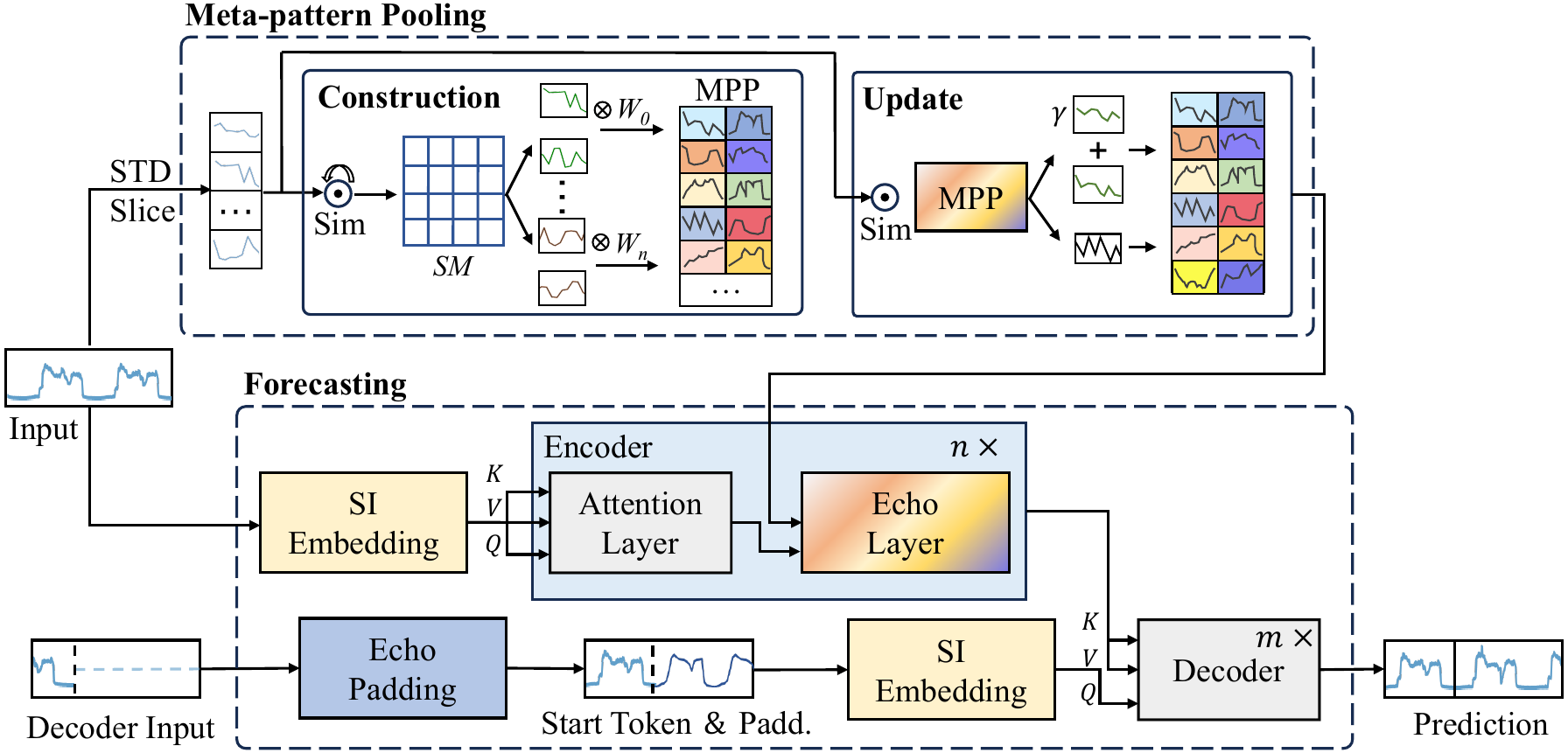}
    \caption{Framework overview of MetaEformer, consisting of two main parts: Meta-pattern Pooling (\S \ref{sec:mpping}) and Forecasting (\S \ref{sec:ff}).}
    \label{fig:framework architecture}
    \Description{Framework overview of MetaEformer}
\end{figure*}

As illustrated in Fig. \ref{fig:framework architecture}, MetaEformer sustains two main processes: Meta-pattern Pooling and Forecasting. The Meta-pattern Pooling involves extracting key components from load series and purifying meta-patterns, alongside maintaining the meta-pattern pool (MPP). The Forecasting elucidates how the proposed Echo Layer and Echo Padding incorporate the MPP to guide the predictive process of the encoder-decoder model. Notably, the two processes are not independent, but tightly coupled. In each computational round, the same input sequences are utilized both for forecasting and for purifying meta-patterns. The updated MPP is then fed into the Forecasting framework's Echo Layer, ensuring seamless integration and continuous refinement.

% Additionally, we develop a novel Static Inject (SI) Embedding to inject supplementary static information pertinent to real-world systems.

% As illustrated in Fig. \ref{fig:framework architecture}, Additionally, we develop a novel Static Inject (SI) Embedding to inject supplementary static information pertinent to real-world systems.

% 如图3所示，整个MetaEformer框架维系两个并行且紧密交互的过程：Meta-pattern Pooling and forecasting。meta-pattern pooling包含对每条负载序列的关键组件提取和元模式识别，对全局元模式池的构建和更新。forecasting则主要介绍了所设计的Echo Layer和Echo Padding是如何利用meta-pattern pool指导encoder-decoder模型的预测过程，除此之外，我们还设计了一种新的Static Inject (SI) Embedding to inject additional static information of real-world systems.

\subsection{Meta-pattern Pooling}\label{sec:mpping}
As aforementioned, meta-patterns can be descriptively defined as \emph{fundamental waveform units that capture the essential periodic and structural components of complex and dynamic time series data.} Beyond this, we emphasize that the operational definition, i.e., \emph{how meta-patterns are purified from general patterns}, is of greater practical value. This involves identifying finer-grained patterns and merging similar ones to achieve more representative meta-patterns.
\subsubsection{\textbf{Seasonal Component Extraction}}
% To learn with the complex temporal patterns in long-term forecasting context, we take the idea of decomposition [1, 33], which can separate the series into trend-cyclical and seasonal parts.
% harnesses the decomposition as an inner block of deep models 这个语句可以参考一下
% 为了应对现实世界中复杂动态的周期模式，将什么信息记录下来能够更加有效地给予后续指导是一个关键问题。我们介绍分解的思想（引用Autoformer），并将其无伤地作为端到端模型的内置组件，来将不同场景下的负载模式分解成趋势与周期部分。具体来说，我们使用移动平均来平滑输入的负载波形并将其视为负载的趋势。对于回望窗口为L的预测任务，这一过程如下：

% To navigate the complex and dynamic periodic patterns in real-world system loads, determining the critical information for future guidance stands as a pivotal challenge. 

% 真实世界系统负载的复杂与动态性的最直接体现在于其周期模式，，因此提取周期模式也是提纯元模式的前置。

The direct manifestation of the complexity and dynamics of system loads lies in their periodic patterns, and thus extracting the periodic patterns serves as the precursor to purifying the meta-patterns. To achieve this, we introduce decomposition~\cite{wu2021autoformer} and seamlessly integrate it as a built-in component. This approach decomposes the load series into trend and seasonal components. Specifically, we use moving averages to smooth the input load fluctuation, treating it as the trend of the load. The process is as follows:
\begin{equation}
\begin{split}
\mathcal{X}^T & =\operatorname{AvgPool}(\operatorname{Padding}(\mathcal{X})),~\mathcal{X} \in \mathbb{R}^{L \times d} \\
\mathcal{X}^S & = \mathcal{X} - \mathcal{X}^T
\end{split}
\end{equation}
where ${\mathcal{X}}^{T}$, $X^S \in \mathbb{R}^{L \times d}$ denote the trend and the seasonal part respectively. We adopt the AvgPool(·) for moving average with the padding operation to keep the series length unchanged. Subsequently, we use STD(·) to describe the whole decompose process.

The trend component indicates the underlying pattern in the data, while the seasonal component captures the periodic cycles. In real-world systems, the trend component is usually simple and stochastic due to human or other interference~\cite{woo2022etsformer}. By employing decomposition, our aim is to extract components with greater representational and reconstructive potential. Therefore, the seasonal part is selected for purifying meta-patterns.

% 我们判断在真实世界系统中，相较于seasonal component, the trend component往往更加简单，broadly divided into Level and Growth~\cite{woo2022etsformer}，另外the trend component也不具备。By employing decomposition, 我们希望寻找到更具代表性和重构潜力的组件以构建元模式池，因此the trend part is chosen to be eliminated，enabling the subsequent process to concentrate more on identifying and memorizing the seasonal parts, which are more complex and dynamic.

% The trend component represents the underlying trend in the data, seasonal component represents the periodic cycles. 相对于周期，趋势更加平稳与简单，大致分为：Level and Growth（引用ETS）。采用分解的操作能够消除平稳的趋势部分，使得XX memory能够更加专注于识别与记忆更具复杂性与动态性的周期部分，从而给复杂动态的现实世界负载预测提供强有力的指导。

%%%%%%%%%%%%%%%%%%%%%%%%%%%%%重构Construction部分%%%%%%%%%%%%%%%%%%%%%%%%%%%%%%%%%%%%%
\subsubsection{\textbf{Meta-pattern Pool Construction}} Although decomposition helps extract periodic patterns of load series, these isolated patterns are not yet quintessential meta-patterns. 
% 需要对periodic patterns进一步提纯，MetaEformer通过维护一个精心设计的MPP来实现这一目的，这个过程包含构建初始的MPP和后续围绕它的更新。
To refine them into distinct, highly representative meta-patterns, further purification is needed. MetaEformer achieves this by maintaining a meticulously designed MPP, involving construction and subsequent updates.

%自动化小波的概念 我们要在这一章节体现

% 如图3所示，MPP的构建发生在模型训练的第一个batch输入阶段，初步构建后的MPP将被MetaEformer的预测过程所利用，并可以在之后的训练与推理阶段得到不断的更新。Once established, the MPP is leveraged by the forecasting process and subsequently updated during ongoing model training and inference.

% MPP的整个构建过程被形式化为算法1，具体来说，MetaEformer会维持一个维度为$P \times s$的MPP$\matcal{P}$（P的具体设置与数据集相关，见实验章节），其初始被mask值0所填充。之后，第一批次的时序数据$X$将经过STD与Slice得到细粒度的周期波形矩阵$W\in \mathbb{R}^{(BL/s) \times s$，通过对$W$进行行间的Sim操作，我们可以得到波形间的相似度矩阵$SM$.

As illustrated in Fig. \ref{fig:framework architecture}, the MPP is initialized with the \textbf{first training batch}, and the whole construction process is formalized in Algorithm \ref{alg:metapattern_pool}. Specifically, the MPP ($\mathcal{P}$) with dimensions $P \times s$ (where the MPP size $P$ is scenario-dependent and discussed in the experiment section) is initially filled with a masking value of zero. Subsequently, the initial batch (with batch size $B$) of time series data $X$ undergoes STD(·) to extract the periodic patterns of loads.

\begin{figure}[tbp]
    \centering
    \includegraphics[width=\columnwidth]{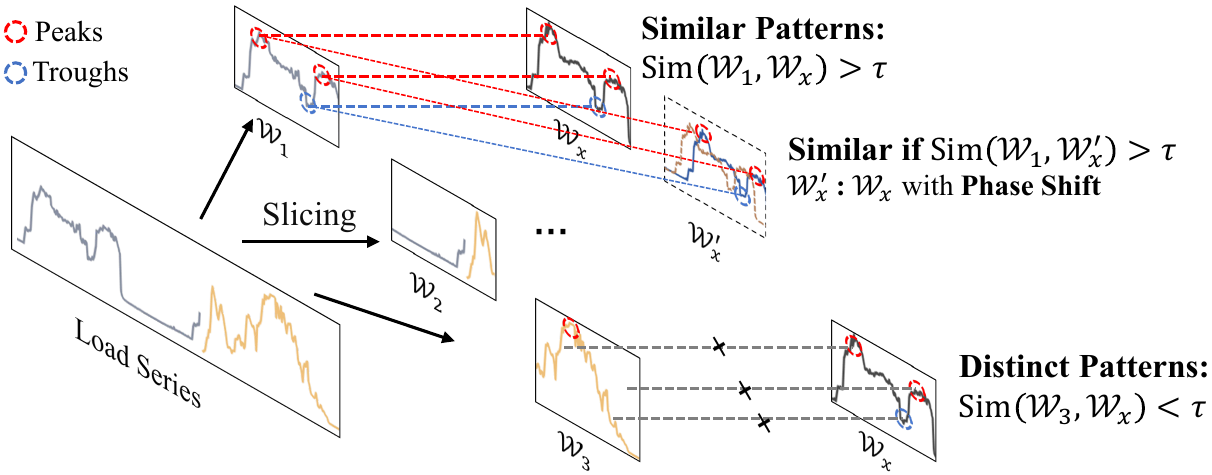}
    \caption{Waveform similarity measurement.}
    \label{fig:sim}
    \Description{Diagram of waveform similarity measurement.}
    % \vspace{-0.8cm}
\end{figure}

In addition to STD(·), inspired by the patch operation in the image domain~\cite{dosovitskiy2020image}, we introduce the Slice(·) operation to yield a more fine-grained waveform matrix $W \in \mathbb{R}^{(BL/s) \times s}$ (where $s$ is the waveform length). While a single time series may contain multiple waveforms, especially in complex and dynamic scenarios, as illustrated in Fig. \ref{fig:sim}, considering a time series with entity switching might encompass three or more patterns. Treating the entire sequence as a whole could lead to the loss of these patterns. 

% As Fig. \ref{fig:sim} shows, the original time series $\mathcal{X}_t$ (with length $L$) will be firstly sliced into $L/s$ waveforms of length $s$, after which these waveforms will be regarded as the basic computational units to compute the similarity among other waveforms.

% 如算法1中的第4行所述，MPP construction的一个重要步骤是衡量waveform之间的相似性，并根据这种相似性判断哪些waveform应当被融合为一种meta-pattern而哪些应该被视为独立的。这一过程应当满足准则：1）相似的波形被最大程度的发现和融合, 以减少冗余的储存和计算；2）异构的波形被最大程度的隔离, 以便尽可能利用信息, 应对复杂动态场景。此外，作为神经网络内部的计算单元，这一过程应当足够简单高效，以避免过高的计算复杂度，因此诸如DTW的带有warping相似性计算方法(计算复杂度为O(s^2))，和高复杂度的的聚类方法都不再适用。

After extracting the waveform matrix, a crucial step in MPP construction is to measure the similarity between patterns and determine which patterns should be merged into a single meta-pattern. This process should adhere to the following principles: 

\textbf{1)} Similar patterns should be maximally identified and merged to reduce redundant storage and computation. 

\textbf{2)} Heterogeneous patterns should be maximally separated to fully exploit information and adapt complex dynamic scenarios. 

Additionally, as a computational unit within the neural network, this process should be simple and efficient to avoid excessive computational complexity. Therefore, methods with warping-based similarity calculations, such as DTW (with complexity $O(s^2)$), and high-complexity clustering methods, are not suitable.

% 描述咱们的方法。
\textbf{Waveform Similarity Measurement.} Inspired by vector dot product similarity, we develop an alignment enhancement method $\text{Sim}(\cdot, \cdot)$ for measuring pattern similarity. Specifically, we approach from a waveform perspective, determining the similarity between patterns by aligning their fluctuations and augmenting the connection among similar patterns. The similarity between two waveforms $\mathcal{W}_1$ and $\mathcal{W}_2$ in the waveform matrix $W$ is computed as follows:  
\begin{equation}
\text{Sim}(\mathcal{W}_1, \mathcal{W}_2) = \sum_{i=1}^{n} w_{1i} \cdot w_{2i}
\end{equation}where $w_{xi}$ denote the $i$-th value of $\mathcal{W}_x$. This method suggests that congruent peaks will lead to a higher $\text{Sim}$, facilitating pattern matching, while disparate patterns result in lower $\text{Sim}$ due to peak and trough misalignment. 

Sim enhances interpretability and is computationally efficient, making it ideal for end-to-end models. Moreover, due to the incorporating of standardization, the method mitigates amplitude variations, ensuring the similarity measurement focuses on the shape of waveforms rather than absolute high values. 
% 这一方法从向量的角度看同样make sense,由于标准化的存在，其可以近似视作两个时序向量之间的余弦相似度，即\text{cosSim}(\mathcal{W}_1, \mathcal{W}_2)=\frac{\mathcal{W}_1\cdot \mathcal{W}_2}{||\mathcal{W}_1||||\mathcal{W}_2||}_{(standardised)}=\text{Sim}(\mathcal{W}_1, \mathcal{W}_2)。This waveform similarity measurement enhances interpretability and avoids high computational expense, ideal for end-to-end models.
% Additionally, due to standardization, this method approximates the cosine similarity between two time series vectors. That is,
% \begin{equation}
% \text{cosSim}(\mathcal{W}_1, \mathcal{W}_2)=\frac{\mathcal{W}_1 \cdot \mathcal{W}_2}{||\mathcal{W}_1|| ||\mathcal{W}2||}_{(standardized)} \approx \text{Sim}(\mathcal{W}_1, \mathcal{W}_2)
% \end{equation}
% \sy{In addition to direct entity switching, complex dynamic scenarios also involve issues like phase drift and variable periods of the same entity over time. This alignment-based method effectively alleviates these problems by recognizing slight phase drift and variable periods as similar patterns with high similarity scores, while significant drifts and periods are identified as distinct patterns with lower scores. Moreover, by incorporating standardization, the method mitigates amplitude variations, ensuring the similarity measurement focuses on the shape of waveforms rather than absolute values.}

After performing Sim between rows of $W$, we obtain the waveform similarity matrix $SM$ ($SM$ is structured as an upper triangular matrix to avoid redundant computations) as follows:
\begin{equation}
SM_{ij} = 
\begin{cases} 
\text{Sim}(W_i, W_j) & \text{for } i < j, \\
0 & \text{for } i \geq j.
\end{cases}
\end{equation}

% 从时序预测的角度来看，相位本身也记录了关键的可预测信息，因此Sim的计算会保留相位偏移所带来的影响，作为后续meta-pattern所蕴含特征的一部分。但值得注意的是，只有较大偏移，伴有较低的Sim会被视为从属不同的meta-pattern，而轻微的偏移会仍会被视为同种meta-pattern，这之间的区分将会通过接下来的阈值机制实现。
For time series forecasting task, \emph{phase shifts} ($\mathcal{W}'_x$ in Fig. \ref{fig:sim}) contain critical predictive information. Therefore, the alignment based Sim retains the impact of phase shifts as part of the meta-pattern features, rather than eliminating it. Significant phase shifts with low Sim values are considered different meta-patterns, while slight phase shifts are considered the same meta-pattern. This distinction is achieved through the purification operation described below.

% 仅靠Sim仍无法满足前文所属的准则P1和P2,因为Sim仅提供了对两个波形间相似性的衡量，但无法准确判断众多波形的聚合或独立，为此我们提出一种基于阈值判断的快速聚合方法。
\textbf{Threshold-based Purification.} Sim measures waveform similarity but does not determine aggregation, making it insufficient to satisfy the principles of pattern clustering and independence. To address this, we introduce a threshold-based purification method. Given the purification threshold $\tau$, patterns with Sim greater than $\tau$ are treated as identical meta-pattern, while others remain separate. Using the similarity matrix $SM$, we compute the purification threshold $\tau$ as follows:
\begin{equation}
\tau = \mu(SM) + \frac{\alpha P}{|SM|}\sigma(SM)
\end{equation}where $\mu(SM)$ represents the mean of the non-diagonal elements of $SM$, $\sigma(SM)$ denotes the standard deviation. Given the generally larger $P$, $\sigma(SM)$ would significantly impacts $\tau$. To balance this, we scale $\frac{P}{|SM|}$ by a constant factor $\alpha$ (set to 0.5). This rationale behind fixing $\alpha$ while adjusting $P$ aligns with the scenario's complexity, as demonstrated in \S \ref{sec:PS}.

% While $P$ is generally larger, leading to a decisive role of $\sigma(SM)$ in the calculation of $\tau$. To counteract this effect, we scale the $\frac{P}{|SM|}$ by a scaling factor $\alpha$ (set to 0.5), 

% This approach dynamically adapts the meta-pattern purification process to the evolving size of the MPP, ensuring diversity and relevance of MPP.

The formula for $\tau$ is dynamically fluctuate with the MPP size $P$, which is scenario-dependent. More complex system loads require a larger $P$, resulting in a higher $\tau$ and ensuring more distinct patterns are recognized as separate meta-patterns, thus preventing excessive amalgamation.
% 正如算法1中的5到13行所示，在明确SM和$\tau$之后，MetaEformer将为每种模式选择所有Sim大于$\tau$的相似模式，并按照其相似度作为权重融合为一种meta-pattern填入至MPP，直到第一批次数据被遍历完。
As outlined in Algorithm \ref{alg:metapattern_pool}, after specifying $SM$ and $\tau$, MetaEformer selects all similar patterns with Sim value greater than $\tau$ for each pattern (line 8), fusing them into a meta-pattern weighted by their similarity (line 9) and populating the MPP (line 10), until the first batch of data is traversed.

\begin{algorithm}[t]
\caption{\textbf{Meta-pattern Pool Construction}}
\label{alg:metapattern_pool}
\begin{algorithmic}[1]
\item[\textbf{Input:}] the first batch input time series $X$, the waveform length $s$, the MPP size $P$ 
\item[\textbf{Output:}] the meta-pattern pool $\mathcal{P}$
\State Initialize empty meta-pattern pool $\mathcal{P} \in \mathbb{R}^{P \times s}$ with mask 0
\State Extract the seasonal part of $X$ and slice into waveform matrix $W = \text{Slice}(\text{STD}(X)), W \in \mathbb{R}^{(BL/s) \times s}$
\State /* \textcolor{blue}{The calculations are described in Eq. 4 and 5} */
\State Calculate similarity matrix \( SM \) and similarity threshold $\tau$
\For{$i = 1$ to $|SM|$}
    \State Determine max similarity $\nu = \max(SM_i)$
    \If{$\nu > \tau$}
        \State Identify similar waveforms $\mathcal{K} = \{ k \mid SM_{ik} > \tau \}$
        \State Purify weighted pattern $M = (W_{\mathcal{K}}^T \cdot \frac{SM_{i\mathcal{K}}}{\sum(SM_{i\mathcal{K}})})$
        \State Append to MPP $\mathcal{P} \leftarrow M$
    \Else
        \State Append original waveform to MPP $\mathcal{P} \leftarrow W_i$
    \EndIf
\EndFor
\State \textbf{return} $\mathcal{P}$
\end{algorithmic}
\end{algorithm}

\subsubsection{\textbf{Meta-pattern Pool Update}}
% 随着模型的训练，MPP会不断地从新的输入中发现并更新自身存储的元模式。这有两种情况：1. 新输入数据中的部分周期模式与已记忆的元模式类似，则会将类似的周期模式融合到对应的元模式中，不断地优化这一元模式以便增加泛性。2. 输入数据中出现了从未见过的新周期模式，与存储过的元模式截然不同，则将其作为新的类别存储到元模式池中。当然，尽管元模式池仅存储切分后的波形模式，存储成本低，可存储海量的类别，但元模式池依旧存在记忆上限。当达到这一上限后，MMP仅会将新的模式与原有最相关的模式进行融合。这一计算过程如下：

% 初始化构建的元模式池由于只依赖第一个batch的训练数据，无法填充足够的元模式来涵盖未来的多种load series。为此，我们设计了针对MPP的更新机制，使得初始化后的MPP能够根据到来新数据不短迭代更新。

% As the model trains, the MPP continuously discovers and updates its stored meta-patterns from incoming data. 
The initially constructed MPP, relying solely on the first batch of training data, is inherently limited in its capacity to encompass the various future load series. To address this limitation, we propose an MPP update mechanism (unfolded in Appendix \ref{sec:MPP_update}, Algorithm \ref{alg:pool update}), enabling MPP to iteratively evolve and incorporate new data post-initialization. This mechanism ensures that the MPP dynamically adapts, extending its coverage to accommodate the evolving landscape of load series.

The mechanism occurs in two ways: 1) Distinct, novel waveforms are categorized as new meta-patterns in the MPP until its size $P$ is reached. 2) Resembling waveforms from new data are merged into the most similar meta-patterns, enhancing their generality.

% Despite its efficient storage of segmented waveforms allowing numerous categories, the MPP's size $P$ imposes a memory constraint. Once this limit is reached, the MPP selectively merges new patterns with the most relevant existing meta-patterns. 

Notably, the Meta-pattern pooling mechanism can be regarded as an efficient and integrated clustering process, supporting the identification of similar load patterns and of unique load behaviors. Clustering-based approaches are widely recognized for their interpretability, as they organize data into meaningful structures \cite{10.1145/3580305.3599453, Yang2023}. \textit{The proposed mechanism inherits this property while avoiding the high computational complexity and potential information leakage.}

%  In MPP, this property ensures that load patterns are systematically grouped based on shared characteristics, providing a clear, data-driven rationale for how meta-patterns evolve.

% 不同于一众基于聚类的池化方法，MPP连同其创建和更新过程，被设计为整个预测框架中的关键组件，而不是预先构建的模型外部输入。这种设计高度契合了时序预测任务的流式输入输出过程，避免了额外的训练成本和潜在的信息泄露问题。
% However, distinct from clustering-based pooling methods \cite{10.1145/3580305.3599453, Yang2023}, the MPP along with its construction and update processes are integral to the framework rather than pre-constructed external inputs. This design aligns with the streaming input-output nature of time series forecasting tasks, avoiding pre-processing and information leakage.

% 值得注意的是，MPP的构建与更新可以被视作一种高效且集成式的聚类学习过程，这一过程支持对相似负载的挖掘以及对独特行为的发现

\subsection{Forecasting Framework}\label{sec:ff}
% 我们的padding是很有价值的 这里和后续的实验都可以重点说一下。
% 简单提一下start token 但是之前的start token的后半部分都是简单的均值和0值，没有什么可解释性 理论上 换成什么值都可以 但是我们的padding 换一个说法，是从元模式池提取信息 更具可解释性

% 在这个部分，我们将介绍专门针对现实应用中复杂性与动态性挑战的MetaEformer模型。MetaEformer模型主要由多个encoder和decoder组成，辅以SA Embedding模块，Echo Padding模块和Echo Layer。MetaEformer的输入包括时序序列本身, 有关序列产生的静态上下文信息和时序mark。MetaEformer采用start token，此外为DMS模型，避免IMS方法的误差累积。

The MetaEformer's forecasting framework is built on transformer-based predictor, primarily consists of multiple encoders and decoders, supplemented by the Static Inject (SI) Embedding module, Echo Layer and Echo Padding module, as shown in Fig.~\ref{fig:framework architecture}. The inputs for Forecasting framework include the time series data (with temporal marks), static context information related to the series generation, and MPP. MetaEformer utilizes a start token and serves as a direct multi-step (DMS) forecasting, avoiding the error accumulation associated with the iterated multi-step (IMS) approach~\cite{chevillon2007direct}.

\subsubsection{\textbf{SI Embedding}}
% 正如INtro部分所讨论到的，现实世界中的负载模式会随着不同的实体与周围环境的变化而变化。以这一特性较为显著的边缘云应用为例，边缘云设备具有显著的异构性，应用所部署的硬件环境对于其负载模式具有一定的影响。因此，在本文中超越经典Transformer提出的Position Embedding和Informer提出的Temporal Embedding，我们设计了Static Inject embedding将额外的静态信息注入。
% 值得注意地是，我们与其他传统地将静态信息叠加在模型输入的方法不同，我们更进一步地通过两个线性层将时序Embedding的信息与静态信息Embedding的信息进行融合，赋予模型自主选择与融合Embedding特征的能力，进而在高维空间中学习静态信息与负载模式之间的关联，因此在类似的外界环境下模型能够更加的鲁棒，应对实体切换与新实体的挑战。模型SI Embedding的计算过程如下：

In real-world systems, the load varies with the environment \cite{LIM20211748}. Taking edge cloud applications for instance, where devices exhibit significant heterogeneity. The hardware environment in which applications are deployed can influence their load patterns. Therefore, going beyond the classical Transformer's Position Embedding~\cite{NIPS2017_3f5ee243} and the Informer's Temporal Embedding~\cite{Zhou2021}, we design the SI Embedding to inject additional static information.

Differing from the traditional approach that simply stacking static context onto model inputs~\cite{LIM20211748}, SI Embedding autonomously merges the temporal and static information through two adaptive linear layers. Consequently, the model can learn the relationship between static information and load patterns in a high-dimensional space, enhancing robustness in different external environments and contributing to overcoming challenges posed by entity changes and new entities. The computation of SI Embedding is as follows:
\begin{equation}
\begin{split}
{E}^{SI} &= \text{Linear}_1(\mathcal{S}) \\
{E}^X &= \text{Token}(X)+\text{Temporal}(X)+\text{Position}(X) \\
{E}^F&=\text{Softmax}(\text{Linear}_2(E^X))\cdot{E}^{SI}\\
{E}^X &= \text{LayerNorm}({E}^X + \text{Dropout}(E^F))
\end{split}
\end{equation}
% 其中S代表静态信息，X代表输入的时序序列。Token, Temporal, Position 分别表示TokenEmbedding, TemporalEmbedding和PositionEmbedding。${E}_F$表示最终融合出的Embedding。
where $\mathcal{S}$ represents static context and $X$ represents the input time series. $\text{Token}$, $\text{Temporal}$, $\text{Position}$ respectively signify TokenEmbedding, TemporalEmbedding and PositionEmbedding. ${E}^F$ denotes the final fused Embedding.

\subsubsection{\textbf{Echo Layer}}
% encoder 由两个部分组成，分别是Attetion Layer 和 Echo Layer。其中Attetion Layer为Full Attention。 Echo Layer被设计用来增强encoder的能力，可以根据Attation Layer输出结果的周期模式，从MPP中回想类似的模式。假设有n个encoder，则第i个encoder的计算过程如下:
The encoder of MetaEformer consists of three parts: the full Attention Layer, Meta-pattern Pool ($\mathcal{P}$) and the Echo Layer, which is formalized as follows: 
\begin{align}
% {E}^F & = \text{SIE}(X^{en}) \\
\mathcal{P} &= \text{Meta-pattern Pooling}(X) \\
% \mathcal{P}_{new} & = \text{Update}\left(X^{en}, \mathcal{P}\right)\\
A^{en}&=\text{Attention}({E}^F) \\
O^{en} &=\text{Echo\ Layer}\left(A^{en}, \mathcal{P} \right)
\end{align}
where Meta-pattern Pooling can be the MPP construction when $X$ is the first batch of input series or the MPP update when $X$ are subsequent inputs. $A^{en}$ is the result of the Attention Layer, and $O^{en}$ is the output of the encoder.

\begin{figure}[tbp]
    \centering
    \includegraphics[width=0.8\columnwidth]{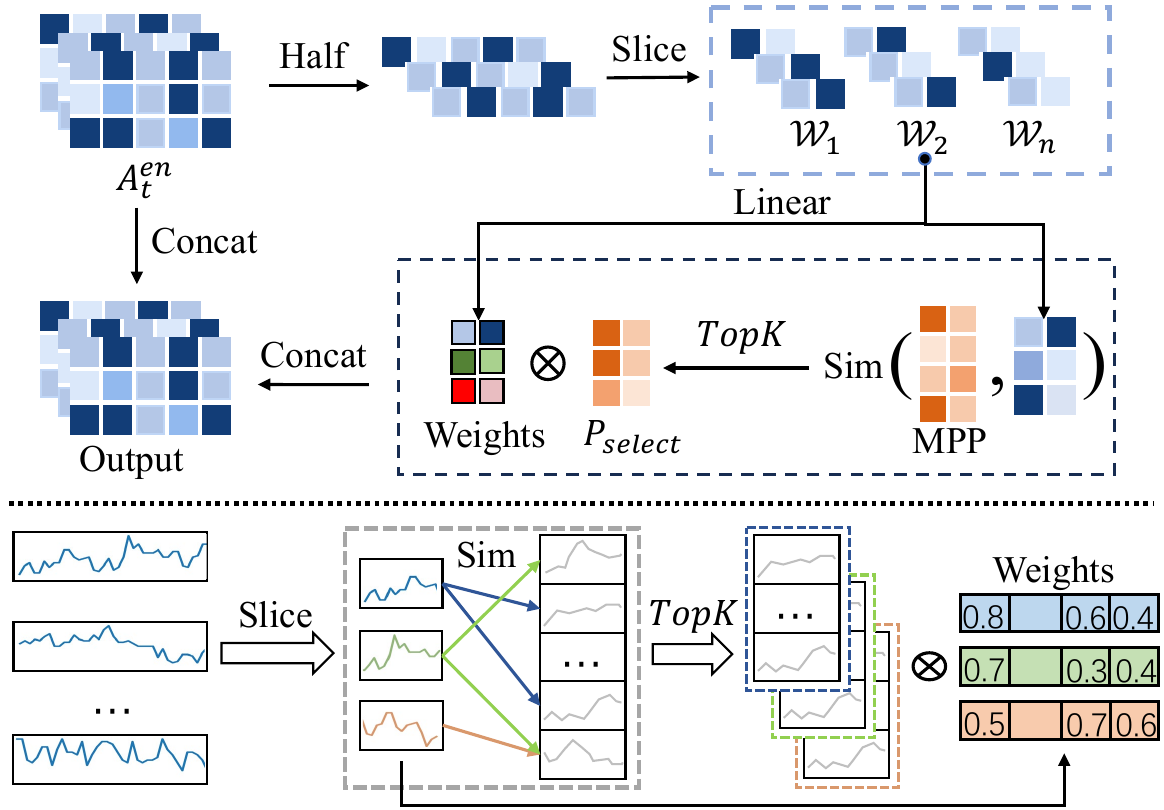}
    \caption{Echo Layer.}
    \label{fig:Echo Layer}
    \Description{Echo Layer.}
    \vspace{-0.5cm}
\end{figure}

As the core of MetaEformer, the Echo Layer (as shown in Fig. \ref{fig:Echo Layer}) is designed to enhance encoder's capabilities by echoing back the most similar meta-patterns, thus focusing the Attention function on the inherent connections and differences in patterns. Specifically, the Echo Layer implements two main operations: deconstructing load series and reconstructing load patterns. The deconstructing operation is to identify the input load series into Top-K most similar meta-patterns by leveraging MPP. The detailed process is as follows:
\begin{align}
W_1, \cdots, W_n &=\text{Slice}_{(2)}(A^{en}[:,:,\frac{1}{2}d_{model}:]), n=L/s \\
\mathcal{P}_{select}^{i}&=\text{Top-K}\left(\text{Sim}(\text{Linear}(W_i), \mathcal{P}) \right)
\end{align}where $\text{Slice}_{(2)}(\cdot)$ denotes the Slice in the second temporal-dimension of $A^{en}$, $W_i \in \mathbb{R}^{B\times s \times \frac{1}{2}d_{model}}$ is the waveform obtained by slicing from $A^{en}$. After that, each waveform $W_i$ is used to Sim with the MPP to identify the $K$ most similar patterns, denoted as $\mathcal{P}_{select}^{i} \in \mathbb{R}^{K\times s}$, Linear represents a dimension reduction operation in order to align the dimensions of $W_i$ with $\mathcal{P}$. Note that we use the last $\frac{1}{2}$ features of $A$ as the operation object, while the first $\frac{1}{2}$ features are retained for keeping the original information.

The reconstructing is to reintegrate the identified meta-patterns as new features in load series, which can be formalized as follows:
\begin{align}
O_i &= \text{Softmax}\left(DR(W_i) \right) \cdot (\mathcal{P}_{select}^i)^T \in \mathbb{R}^{B\times s \times K}\label{eq:1} \\
% \uparrow_1    \\
\mathbf{O} & = \text{Concat}\left(DI(O_1,\cdots,O_n)\right) \in \mathbb{R}^{B\times L\times \frac{1}{2}d_{model}} \\
O^{en} &= \text{Concat}(A^{en}[:,:,:\frac{1}{2}d_{model}],\mathbf{O}) \in \mathbb{R}^{B\times L\times d_{model}}
\end{align}
where $DR$ is the dimension reduction Linear to align $W_i$ with $\mathcal{P}_{select}^i$, $DI$ is the dimension increase Linear to recovery the needed dimension. In Eq. \ref{eq:1}, the Echo utilizes $W_i$ to adaptively generate the weights to merge $\mathcal{P}_{select}^{i}$, extracting information from the Top-K similar meta-patterns. After that, $K$ will be the new feature dimension. The information merged from multiple meta-patterns is then concatenated with the first half of $A^{en}$, forming the final output.

% 其中Slice为切分操作，如前文所说，为发现更加细粒度的模式，会将序列切分为若干个子段，如S1，Si,Sn等，n为输入序列除以子段长度，子段长度一般设置为能够被L整除。子段Si会与元模式池相乘以便寻找最相似的K个模$\mathcal{P}_select^{i}$。此后会根据Si选择合适的权重将$\mathcal{P}_select^{i}$进行融合，提取元模式的相关信息。多个子段融合后的信息将会拼在一起并与A的前一半相拼接，作为最终的输出。

\subsubsection{\textbf{Decoder}}
% decoder 部分包含Echo Padding, SI Embedding 和 堆叠的M个 decoder layer。假设encoder输出的潜在变量O和decoder input X，decoder部分可以被形式化为如下表达：
The decoder comprises Echo Padding, SI Embedding, and a stack of M decoder layers. Assuming the latent variables $\mathcal{O}^{en}$ as the output of the encoder and $\mathcal{X}^{de}$ as the decoder input, the decoder can be formalized as follows:
\begin{equation}
\begin{split}
    X^{de} & = \text{Echo\ Padding}(X^{de}) \\
    {E}^{de} & = \text{SI Embedding}(X^{de}) \\
    O^{de} & = \text{decoder}({E}^{de})
\end{split}
\end{equation}

\subsubsection{\textbf{Echo Padding}}
% 为了实现DMS预测，不同于以往使用占位符填补预测长度的做法，我们从元模式池中提取相关的元模式进行融合来填充预测。此前的占位符，有些模型是输入序列的均值，有些则直接为零，这显然缺乏可解释性，也无法为decoder提供更多的信息，从而提升模型性能。基于前文设计的元模式池，Echo Padding重复利用这一设计的强大之处，可以依据输入的时序序列，回忆与当前序列最相关的元模式并将其进行融合，以便生成基础的周期模式当做预测序列的填充。就近的序列往往能给预测提供较大比重的信息，这一设计可以使decoder在这一基础周期模式之上进行预测序列的重构，提升模型的信息利用率与预测性能。

Decoder inputs padding is a key component for transformer-based models to achieve DMS prediction \cite{Zhou2021}. The previous practice use mean or simply zeros placeholders to fill the prediction length, lack interpretability and fail to provide additional information, thus limiting model's performance. To address this, we propose the Echo Padding to reuse the powerful MPP and fuse the relevant meta-patterns for filling the decoder inputs:
\begin{align}
    X^{EP} &= \text{Contact}_{i=1}^n\left(\text{Softmax}\left(DR(W_i) \right)\odot(\mathcal{P}_{select}^i)^T\right) \in \mathbb{R}^{B\times L_y \times d} \nonumber \\
    X^{de} &= \text{Concat}\left(X[:,L_{token}:,:], X^{EP}\right)\in \mathbb{R}^{B\times (L_{token} + L_y)\times d} \nonumber
\end{align}
where $\odot$ is the matrix multiplication form of Eq. 12, $L_{token}$ is the token length. This design allows the decoder to reconstruct the predictive series based on meta-patterns, enhancing the model's information utilization and predictive performance.

\subsection{Negligible Additional Complexity}\label{sec:additioanl_com}
% Autoformer的图来论证Transformer在非超长期方面的性能有效性。之后这一块直接给结论，原生trans的复杂度是多少，我们是多少，增加的倍数是多少，最后我们比较了性能与时间开销的平衡如下图所示。显示出我们The experimental results below also demonstrate a significant increase in the model's accuracy and robustness for complex dynamic system load forecasting with minimal extra computational cost.

% MetaEformer所引入的Meta-pattern Pooling与Echo机制实现了对meta-pattern的识别与利用，确保其在复杂动态系统下的跨场景有效性。虽然其在传统时序Transformer上引入了额外的机制，但MetaEformer谨慎地看待MPP的构建与更新，同时Echo对top-k meta-pattern的选择都使其新增的计算复杂度相当有限。具体来说：
% The Meta-pattern Pooling and Echo mechanisms enable the purification and leveraging of meta-patterns, ensuring MetaEformer's cross-scenario effectiveness. 
While the Meta-pattern Pooling and Echo mechanisms these mechanisms add components to the time series Transformer, MetaEformer carefully manages the construction and updating of the MPP, and the Echo Layer selectively employs the Top-K meta-patterns, thus keeping the additional computational complexity relatively low.

% 我们分析了MetaEformer的计算复杂度并将与传统时序Transformer的计算复杂度进行了对比（具体过程见附录）。总结来说，由于MetaEformer谨慎的……其额外机制只引入了$O(B(\frac{Bs}{E}+\frac{LP}{M}+sK))$的计算复杂度，而必要的attention与feed-forward计算的复杂度为O(B N (L^2d + L d^2))，后者是前者的\frac{N(Ld+d^2)}{K}倍。 With common deep learning task settings, this scale can be hundreds of times larger. Therefore, we assert that the additional computational complexity from MetaEformer is \textbf{negligible} in contrast to the native complexity of Transformer-based models.
We analyze the complexity of MetaEformer and compare it with that of the traditional time series Transformer (detailed in the Appendix \ref{sec:complexity}). In summary, due to MetaEformer's careful computation management, the complexity of additional mechanisms could be hundreds of times ($\frac{N(Ld+d^2)}{K}$) \textbf{lower} compared to the original attention and feed-forward computations, rendering it \textbf{negligible}.

Additionally, it is noteworthy that MetaEformer remains competitive in computational efficiency, even with full-attention mechanism. As demonstrated in \cite{wu2021autoformer}, while full-attention mechanisms can be costly for very long sequence predictions, they are more efficient than other attention-variant models for moderate lengths, maintaining the lowest running time until output lengths exceed 2048. This is much larger than the typical settings for system load predictions that we focus on. The conclusion is also supported by the computational efficiency experiments in \S \ref{sec:CE}.

\begin{table*}[htbp]
\setlength{\tabcolsep}{2.5pt}
  \centering
  \caption{Load forecasting performance under various system scenarios. A lower MSE or MAE indicates a better prediction.}
    \begin{tabular}{c|cccccccc|cccccc|cc}
    \hline
    \multicolumn{1}{c|}{\multirow{2}[2]{*}{Datasets}} & \multicolumn{8}{c|}{Cloud}                                    & \multicolumn{6}{c|}{Power}                    & \multicolumn{2}{c}{Traffic} \\
          & \multicolumn{2}{c}{CBW} & \multicolumn{2}{c}{ECW} & \multicolumn{2}{c}{New APP} & \multicolumn{2}{c|}{Switch} & \multicolumn{2}{c}{ECL\_32} & \multicolumn{2}{c}{ECL\_57 } & \multicolumn{2}{c|}{ECL\_270 } & \multicolumn{2}{c}{Traffic\_852} \\
    \hline
    Metrics & MSE & MAE & MSE & MAE & MSE & MAE & MSE & MAE & MSE & MAE & MSE & MAE & MSE & MAE & MSE & MAE \\
    \hline
    \textbf{MetaEformer} & \textbf{0.103} & \textbf{0.130 } & \textbf{0.066} & \textbf{0.137}  & \textbf{0.030 } & \underline{0.118}  & \textbf{0.055} & 0.155  & \textbf{0.029 } & 0.130  & \textbf{0.082}  & \textbf{0.170}  & \underline{0.042}  & \underline{0.149}  & \textbf{0.083 } & \textbf{0.132} \\
    UniTime & 0.202  & 0.189  & 0.105  & 0.183  & 0.039  & 0.127  & 0.058 & 0.149 & 0.037  & 0.150  & 0.127  & 0.214  & 0.152  & 0.303  & 0.117  & 0.167  \\
    PaiFilter & 0.179  & 0.172  & 0.079  & 0.139  & \underline{0.032}  & \textbf{0.112 } & 0.059 & 0.153 & 0.035  & 0.136  & 0.103  & 0.193  & \textbf{0.038 } & \textbf{0.143 } & 0.099  & 0.173  \\
    FreTS & 0.164  & 0.156  & 0.088  & 0.159  & 0.044  & 0.152  & 0.057 & 0.148 & 0.047  & 0.158  & 0.087  & 0.214  & 0.044  & 0.150  & 0.181  & 0.219  \\
    TSMixer & 0.174  & 0.172  & 0.099  & 0.171  & 0.045  & 0.154  & 0.057 & 0.157 & 0.036  & 0.138  & 0.123  & 0.212  & 0.048  & 0.154  & 0.156  & 0.188  \\
    Crossformer & \underline{0.107}  & 0.137  & 0.114  & 0.190  & 0.040  & 0.140  & 0.064 & 0.163 & 0.046  & 0.152  & 0.088  & 0.212  & 0.044  & 0.151  & 0.167  & 0.216  \\
    TimesNet & 0.155  & 0.156  & 0.081  & 0.143  & 0.046  & 0.140  & 0.066 & 0.160 & \underline{0.031}  & \underline{0.128}  & 0.099  & \underline{0.186}  & 0.043  & 0.149  & 0.091  & 0.152  \\
    \midrule
    DLinear & 0.114  & \underline{0.131}  & 0.123  & 0.202  & 0.039  & 0.133  & 0.067  & 0.162  & 0.062  & 0.199  & 0.136  & 0.226  & 0.068  & 0.198  & 0.098  & 0.177  \\
    Quatformer & 0.212  & 0.202  & 0.192  & 0.272  & 0.054  & 0.158  & 0.059  & \underline{0.143}  & 0.168  & 0.329  & 0.160  & 0.263  & 0.077  & 0.207  & 0.106  & 0.200  \\
    FEDformer & 0.249  & 0.237  & 0.160  & 0.257  & 0.047  & 0.152  & \underline{0.056}  & \textbf{0.141 } & 0.184  & 0.346  & 0.126  & 0.231  & 0.060  & 0.181  & 0.107  & 0.198  \\
    Autoformer & 0.164  & 0.193  & 0.079  & 0.146  & 0.111  & 0.229  & 0.158  & 0.270  & 0.166  & 0.320  & 0.141  & 0.248  & 0.064  & 0.188  & 0.138  & 0.241  \\
    \midrule
    SAN & 0.176  & 0.182  & 0.178  & 0.266  & 25.913  & 3.696  & 1.688  & 0.815  & 0.049  & \textbf{0.117 } & 0.116  & 0.209  & 0.044  & 0.150  & 0.098  & 0.171  \\
    Koopa & 0.197  & 0.192  & 0.093  & 0.162  & 0.058  & 0.163  & 0.072  & 0.176  & 0.037  & 0.139  & \underline{0.086}  & 0.204  & 0.047  & 0.156  & 0.166  & 0.199  \\
    Non-sta. & 0.203  & 0.183  & 0.089  & 0.175  & 0.063  & 0.175  & 0.068  & 0.171  & 0.043  & 0.161  & 0.132  & 0.224  & 0.047  & 0.161  & \underline{0.090}  & \underline{0.133}  \\
    \midrule
    DynEformer & 0.124  & 0.161  & \underline{0.067}  & \underline{0.141}  & 0.090  & 0.207  & 0.067  & 0.169  & 0.063  & 0.184  & 0.138  & 0.239  & 0.089  & 0.224  & 0.144  & 0.257  \\
    VaRDE-L & 0.541  & 0.455  & 0.150  & 0.218  & 0.556  & 0.502  & 0.270  & 0.369  & 0.525  & 0.555  & 0.379  & 0.366  & 0.165  & 0.285  & 0.532  & 0.554  \\
    \hline
    \end{tabular}%
   \label{tab:main_result}
\end{table*}%

\section{Experiments}\label{Exp}

\textbf{Datasets.} To validate MetaEformer across diverse system scenarios, we collect datasets from power grids, cloud platforms, and traffic systems. However, these datasets vary in their representation of complexity and dynamics. To magnify the challenges, we employ the Augmented Dickey-Fuller (ADF) test \cite{elliott1992efficient} to evaluate the datasets' non-stationarity, a key indicator of their dynamic nature. Higher ADF statistics and P-values signify greater variability. For multi-series datasets, ADF is computed per series and averaged. We apply a 99\% confidence interval to determine non-stationarity. Complete ADF results are reported in Appendix \ref{sec:ADF}.

Based on the ADF results, we select three non-stationary time series from power datasets~\cite{Zhou2021, li2019enhancing}: ECL\_\{32, 57, 270\}, originating from ECL. For cloud datasets, following~\cite{10.1145/3580305.3599453}, we opt for the ECW dataset, particularly with Switch and New APP datasets for only model testing, highlighting the model's efficacy with frequent changes and new entities. Furthermore, we enrich this scenario with the CBW (Cloud Bandwidth Workload) dataset, a network load collection from a commercial cloud platform, capturing real-world, highly dynamic scenarios. Among three traffic datasets~\cite{chen2001freeway, li2018diffusion}, we select the series with the highest ADF value, which demonstrate notable dynamic shifts in its latter half, as illustrated in Fig.~\ref{fig:scence_analysis}.

\textbf{Baselines.} Fifteen baselines are compared, including six generic forecasting models: UniTime~\cite{liu2024unitime}, PaiFilter~\cite{yi2024filternet}, FreTS~\cite{yi2024frequency}, TSMixer \cite{ekambaram2023tsmixer}, Crossformer~\cite{zhang2023crossformer}, and TimesNet~\cite{wu2022timesnet}; four decomposition-based models: DLinear \cite{zeng2023transformers}, Quatformer \cite{10.1145/3534678.3539234}, FEDformer \cite{zhou2022fedformer} and Autoformer \cite{wu2021autoformer}; three anti-non-stationary models: SAN \cite{liu2023adaptive}, Koopa \cite{liu2023koopa}, and Non-stationary \cite{liu2022non}; two clustering-based models: DynEformer \cite{10.1145/3580305.3599453} and VaRDE-L \cite{Yang2023}.

\subsection{Implementation Details}

All datasets have a sampling interval of 1 hour. The input load series length is $L=48$, and the prediction length is $L_y=24$, aiming to predict the load for the next day. As per the original study guidelines, dataset splits are as follows: the ECW dataset follows a 6:2:2 training, validation, and test ratio; the ECL dataset uses a 7:1:2 split, while the Traffic dataset is divided in an 8:1:1 ratio to capture end-of-dataset shifts without full disclosure. The CBW dataset, specifically collected for this research, adheres to an 8:1:1 division.

For the Meta-pattern Pooling mechanism, the STD decomposition period is fixed at 24, with a slice length $s=16$. The MetaEformer's MPP updates every 50 batches, applying an update rate $\gamma=0.1$. Key parameters, MPP size $P$ and Echo's Top-K are detailed in $\S \ref{sec:PS}$. Models are trained with the ADAM \cite{kingma2014adam} optimizer with an initial learning rate of $10^{-4}$. Batch size is set to 256 and the start token length $L_{token}$ is set to 12. All experiments are repeated three times. \textbf{Metrics}:  MSE $=\frac{1}{n} \sum_{i=1}^n(Y-\hat{Y})^2$ and MAE $=\frac{1}{n} \sum_{i=1}^n|Y-\hat{Y}|$.

\subsection{Main Results}

The principal experiment results are reported in Table \ref{tab:main_result}, from which we can observe that: \textbf{(1)} MetaEformer achieves consistent state-of-the-art performance compared to 15 baselines across 8 load benchmarks under 3 system scenarios. Specifically, against generic models, MetaEformer decreases overall MSE by 24.3\%/ 19.5\%/ 34.2\% in cloud, power, and traffic scenarios, respectively; it surpasses \emph{decomposition-based models} with MSE reductions of 38.3\%/ 51.4\%/ 24.8\% and \emph{anti-non-stationary models} with reductions of 45.6\%/ 24.5\%/ 24.4\%. Last but not least, it outperforms \emph{clustering-based models} by diminishing 51.8\%/ 65.9\%/ 63.4\% MSE. 

(2) MetaEformer exhibits superior adaptability in dynamic system scenarios, with its performance advantage increasing as system dynamics intensify. While the carefully selected dynamic datasets degrade baseline performance (e.g., most clustering-based and gene-ric models see their MSE double from ECW to CBW), MetaEformer effectively mitigates this impact. Its lead over ECW’s second-best model, DynEformer, expands from 1\% to 20\% on CBW.

(3) As state-of-the-art models under different classes, TimesNet, Dlinear, SAN etc. achieve sporadic best/second-best results in Table \ref{tab:main_result}). However, these models' optimal performance is confined to specific scenario and lacks the generalizability that MetaEformer offers across various systems. Beyond accuracy, MetaEformer also provides the efficiency and interpretability essential for real-world industrial forecasting, which are absent in the baseline models.

\subsection{Ablation Study and Model Interpretation} 
As shown in Table~\ref{tab:ablation}, we conduct a series of ablation studies on MetaEformer, with specific datasets for testing each system scenario: CBW-Cloud, ECL\_270-Power, Traffic\_852-Traffic.

\textbf{Meta-pattern Pool.} De-MPP deactivates the MPP component, meaning the meta-patterns are stochastically initialized and no longer updated. The MPP absence leads to an average ascend in MSE and MAE by 54\% and 34\%, respectively. This substantial improvement underscores the necessity of purifying and maintaining meta-patterns, which are crucial for capturing and adapting to the nuanced complexity and dynamics of time series data.

Thanks to the clustering-like paradigm, the Meta-pattern Pooling process, including MPP initialization and updates, is inherently interpretable and can be clearly visualized (detailed in Appendix \ref{sec:inter MPP}), helping operators understand common patterns of full system loads and the model's core focus during real-world deployment.

\textbf{Echo Layer.} De-EL replace the Echo Layer with a linear layer to directly fuse MPP to forecasting, resulting in a 33\% increase in MSE and 21\% in MAE. The Echo Layer's primary function is to adaptively leverage the MPP to reconstruct patterns from deconstructed loads, which is especially effective in tackling the concept drift and few-shots challenges. The results indicate its criticality in maintaining forecasting accuracy in the face of dynamic system behaviors. We further illustrate the Echo Layer in Appendix~\ref{sec:inter Echo}, showing how it leverages the MPP to adaptively drive load forecasting. 

% We also explain and visualize the computation of the Echo Layer in Appendix \ref{sec:inter Echo}, , demonstrating how it leverages the MPP to adaptively drive load forecasting. This interpretability analysis provides insights into the load patterns associated with the predicted entities and offers a clear view of the potential information utilized for padding (Echo Padding).

\textbf{Echo Padding and SI Embedding.} Deactivating Echo Padding (De-EP) and reverting to zero-padding results in a moderate performance decline, with up to a 10\% increase in MSE. This suggests that padding meta-patterns provides more informative context compared to placeholders, particularly in scenarios with frequent few-shot entity introductions (e.g., N. APP). Similarly, replacing the SI Embedding with a direct stacking approach (De-SI) increases MSE by 13\%, spotlighting the utility of integrating temporal and static information via learnable linear layers.

\begin{table}[t]
\setlength{\tabcolsep}{2.5pt}
  \centering
  \caption{Ablation of MetaEformer in various datasets. The ‘Promo.’ indicate the average enhancement in MSE and MAE.}
    \begin{tabular}{cccccccc}
    \toprule
    \multicolumn{2}{c}{Dataset} & CBW   & {E.\_270} & {T.\_852} & { Switch} & {N. APP} & {Promo.} \\
    \midrule
    \multirow{2}[2]{*}{MetaE.} & MSE   & {0.103 } & {0.042 } & {0.083 } & {0.056 } & {0.030 } & \multirow{2}[2]{*}{\textbf{-}} \\
          & MAE   & {0.130 } & {0.149 } & {0.132 } & {0.155 } & {0.118 } &  \\
    \midrule
    \multirow{2}[2]{*}{De-MPP} & MSE   & 0.293  & 0.096  & 0.152  & 0.096  & 0.042  & \textbf{54}\% \\
          & MAE   & 0.262  & 0.211  & 0.174  & 0.235  & 0.149  & \textbf{34}\% \\
    \midrule
    \multirow{2}[2]{*}{De-EL} & MSE   & 0.228  & 0.046  & 0.093  & 0.065  & 0.033  & \textbf{33}\% \\
          & MAE   & 0.257  & 0.159  & 0.147  & 0.180  & 0.127  & \textbf{21}\% \\
    \midrule
    \multirow{2}[2]{*}{De-EP} & MSE   & 0.120  & 0.050  & 0.093  & 0.056  & 0.031  & \textbf{10}\% \\
          & MAE   & 0.157  & 0.167  & 0.146  & 0.156  & 0.120  & \textbf{8}\% \\
    \hline
    \multirow{2}[2]{*}{De-SI} & MSE   & 0.123  & \multicolumn{2}{c}{\multirow{2}[2]{*}{w/o~static}} & 0.060  & 0.035  & \textbf{8\%} \\
          & MAE   & 0.214  & \multicolumn{2}{c}{} & 0.174  & 0.123  & \textbf{13\%} \\
    \hline
    \end{tabular}%
  \label{tab:ablation}%
  \vspace{-0.3cm}
\end{table}%

\subsection{Parameters Sensitivity}\label{sec:PS}

\begin{figure}[ht]
    \centering
    \setlength{\belowcaptionskip}{-0.3cm}
    \includegraphics[width=\columnwidth]{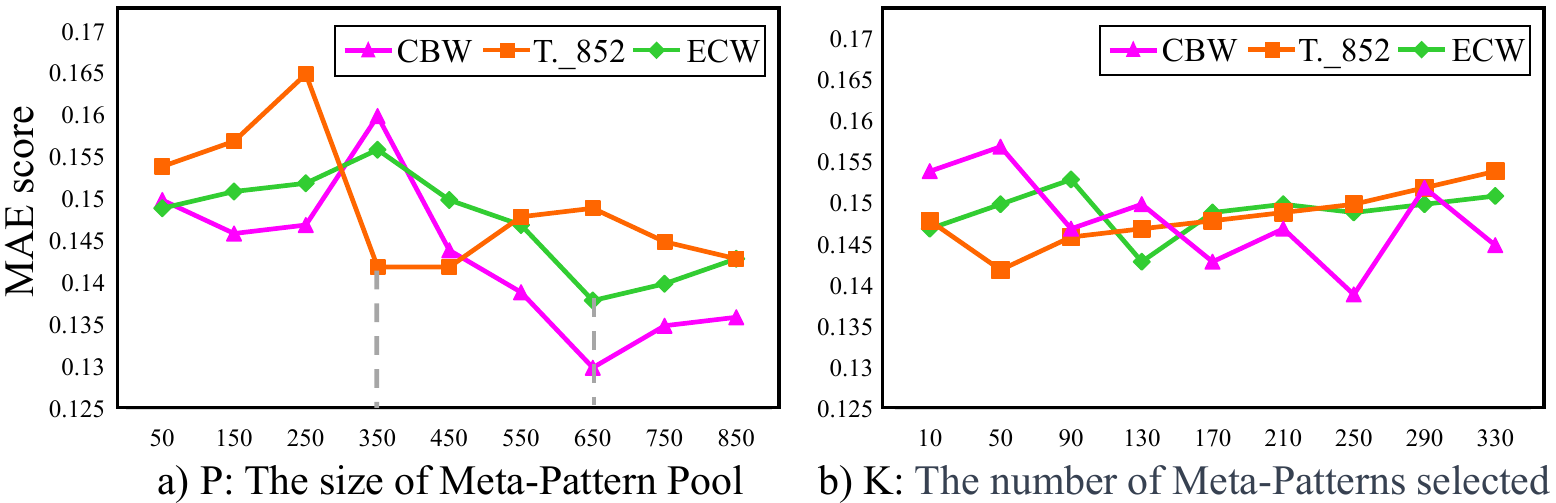}
    \caption{Performance of MetaEformer at different MPP size $P$ and number $K$ of meta-patterns selected by Echo Layer.}
    \label{fig:parameter}
    \Description{Performance of MetaEformer at different MPP size $P$ and number $K$ of meta-patterns selected by Echo Layer.}
\end{figure}

In this section, we explore the sensitivity of two key parameters, 
$P$ and $K$, through experiments on diverse domain datasets. 
$P$ determines the capacity of MPP to sustain meta-patterns, significantly impacting performance and outcomes in various system contexts. As shown in Fig. \ref{fig:parameter}(a), MAE decreases and then increases as $P$ grows from 50 to 850, with the optimal 
$P$ differing by scenario. We observe a positive correlation between this optimal value and the scenario’s
dynamism. Specifically, the most dynamic cloud scenarios exhibit
the largest optimal $P$ values (650 for CBW and ECW), while the
relatively stable traffic scenarios have the smallest $P$ value (350 for Traffic).
$K$ denotes the count of relevant meta-patterns selected from the MPP per Echo. Fig. \ref{fig:parameter}(b) indicates MAE stability for $K$ within 10 to 330. We attribute this stability to the adaptive fusion weights within the Echo, which further amplify or filter effects, making the Echo’s impact relatively stable over a wide $K$ range.

\subsection{Computational Efficiency}\label{sec:CE}

\begin{figure}[ht]
    \centering
    \setlength{\belowcaptionskip}{-0.3cm}
    \includegraphics[width=\columnwidth, height=3.8cm]{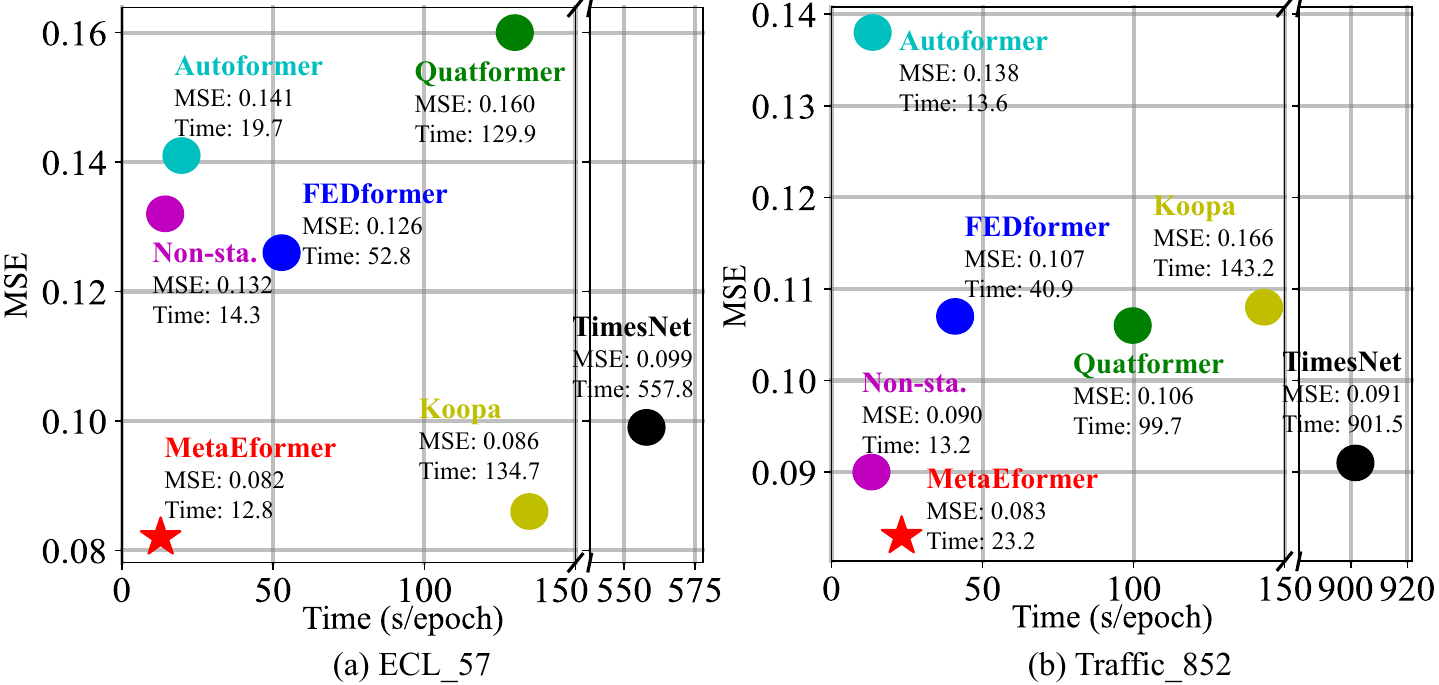}
    \caption{Efficiency of Transformer-based models.}
    \label{fig:time_compare}
    \Description{Model efficiency comparison among Transformer-based models.}
\end{figure}

Fig.~\ref{fig:time_compare} shows the average training time per epoch and the corresponding prediction performance (MSE) for different models on different datasets. MetaEformer shows lower training times with top prediction accuracy, highlighting its superior efficiency and performance. This supports the analysis in \S \ref{sec:additioanl_com}, where the added modules in MetaEformer have negligible time complexity compared to the Transformer. Compared to other Transformer-based models, MetaEformer exhibits competitive time efficiency and the highest predictive performance on both datasets (ECL\_57 and Traffic\_852). This means that MetaEformer can meet the demands of most practical applications, even those with stringent time requirements.

% 模型时间开销对比在Traffic和ECL数据集上

\section{Conclusion}\label{conclusion}
This paper presents MetaEformer, a model built on the novel paradigm of meta-patterns, designed to forecast complex and dynamic loads in various real-world systems, addressing the challenges of complex patterns, concept drift and few-shot. MetaEformer introduces a novel Meta-pattern Pooling mechanism for purifying representative meta-patterns and an Echo mechanism for adaptively deconstructing and reconstructing load patterns. Experiments demonstrate that MetaEformer outperforms fifteen state-of-the-art baselines across diverse datasets, while offering end-to-end efficiency and clear interpretability, making it particularly well-suited for system load forecasting tasks in complex and dynamic environments.

\begin{acks}
We thank the anonymous reviewers and the area chair for their insightful comments. This work is supported by the Beijing-Tianjin-Hebei Basic Research Cooperation Program (F2024201070), the Key Program of the National Natural Science Foundation of China (U23B2049), the Tianjin Natural Science Foundation (23JCYBJC00780), the Tianjin Xinchuang Haihe Lab (22HHXCJC00002), and the National Natural Science Foundation of China (62472311).
\end{acks}

\bibliographystyle{ACM-Reference-Format}
\bibliography{sample-base}

% \newpage
% \balance
\appendix
\section{Appendix}

\subsection{Meta-pattern Pool Update Algorithm}\label{sec:MPP_update}

% Meta-pattern Pool Update的具体流程如算法2所示.
The process of meta-pattern pool update is shown in Algorithm 2.

\begin{algorithm}
\caption{\textbf{Meta-pattern Pool Update}}
\label{alg:pool update}
\begin{algorithmic}[1]
\item[\textbf{Input:}] the input series $X$, waveform length $s$, threshold $\tau$, update rate $\gamma$
\item[\textbf{Output:}]  Updated Meta-pattern pool $\mathcal{P}$
\State Extract waveforms $W = \text{Slice}(\text{STD}(X))$
\State Calculate similarity matrix $SM$ between $W$ and $\mathcal{P}$
\State Find max similarity and corresponding index $(\mathcal{V}_{\max}, \mathcal{I}_{\max}) = \text{Max}(SM, \text{axis}=1)$
\State Filter rows with similarity above threshold $\mathbf{U} \gets \mathcal{V}_{\max} > \tau$
\State $\mathcal{P}[\mathcal{I}_{\max}[\mathbf{U}]] = (1-\gamma)\mathcal{P}[\mathcal{I}_{\max}[\mathbf{U}]] + \gamma W[\mathbf{U}]$

\For{$idx$ in $\{i | \mathbf{U}[i] = \text{false}\}$}
    \State /* \textcolor{blue}{$\mathcal{P}$ has been filled.} */
    \If{$\neg \exists$ row $r$ in $\mathcal{P} : r = \mathbf{0}$}
        \State $\mathcal{P}[\mathcal{I}_{\max}[idx]] = (1-\gamma)\mathcal{P}[\mathcal{I}_{\max}[idx]] + \gamma W[idx]$
    \Else
        \State $\mathcal{P}[r] = W[idx]$
    \EndIf
\EndFor
\State \textbf{return} $\mathcal{P}$
\end{algorithmic}
\end{algorithm}

\subsection{Computational Complexity Analysis}\label{sec:complexity}
% 在这一节中，我们详细给出MetaEformer的计算复杂度，作为Transformer-based models之一，MetaEformer包含了Transformer的核心计算单元，因此我们先说明Transformer模型的计算复杂度，再说明MetaEformer的新增模块。
This section analyzes the computational complexity of MetaEformer. As a Transformer-based model, MetaEformer includes the basic computational units of the Transformer. Therefore, we first outline the complexity of the Transformer, followed by the additional complexity introduced by the new modules in MetaEformer.

\textbf{Transformer Complexity.} For an input length $L$, batch size $B$, and hidden layer dimension $d$, the computational complexity of a Transformer primarily arises from two sources, \textbf{Self-attention:} $O(L^2d)$ and \textbf{Feed-forward layers:} $O(Ld^2)$. If the Transformer has $N$ layers, the total complexity is:
\begin{equation}
O(B N (L^2d + L d^2))
\end{equation}

\textbf{MetaEformer's Modules.} Assuming a slice length $s$, each sequence can be divided into $L/s = n$ slices (usually small and can be neglected), and the size of the MPP is $P$. The additional complexities introduced by MetaEformer can be categorized as follows:

\begin{enumerate}[leftmargin=*]
    \item \textbf{MPP Construction.} The complexity of MPP construction includes three components: (1) Decomposition and slicing, both with time complexity $O(BL)$; (2) Similarity calculation, with complexity $O\left(\left(\frac{BL}{s}\right)^2 \cdot s \right) = O(B^2 s)$; and (3) The operations of finding maximum similarity, checking thresholds, and generating the weighted MPP, which incur a complexity of $O(BN)$. Among these, the similarity calculation is the dominant factor, resulting in an overall complexity of $O(B^2 s)$.

    \item \textbf{MPP Updating.} The primary computational cost of MPP updating lies in the similarity calculation, which has a time complexity of $O\left(B \cdot \frac{L}{s} \cdot P \cdot s \right) = O(B L P)$.

    \item \textbf{Echo Layer.} The complexity of the Echo Layer is mainly determined by two operations: (1) Top-K similarity calculation, with complexity $O\left(n \cdot B \cdot \frac{L}{s} \cdot P \cdot s\right) \approx O(B L P)$; and (2) The Softmax and concatenation steps, with complexity $O(B s K)$.
\end{enumerate}

Since the computational modules are executed serially, the total new complexity introduced by MetaEformer is $O(B^2s) + O(BLP) + O(BLP) + O(BsK) \approx O(B(Bs + LP + sK))$. Note that MetaEformer does not perform MPP construction and updating at every training iteration. Specifically, MPP construction only occurs during the first training batch, while updating frequency is generally set to $\frac{1}{M}$ of training iterations. This reduces the added complexity to:
$$O(B(\frac{Bs}{E}+\frac{LP}{M}+sK))$$
where E is the total number of iterations required for training (epoch*batches per epoch), $M$ is the frequency of performing MPP updates ($M=50$ in our setup). Ultimately, the complexity of the native components of Transformer exceeds the new components introduced by MetaEformer by an order of:
$$\frac{N(Ld+d^2)}{K}$$
with common deep learning task settings, this scale can be hundreds of times large. Therefore, we assert that the additional computational complexity from MetaEformer is \textbf{negligible} and ignorable.

\subsection{ADF Test Results}\label{sec:ADF}
In Table \ref{tab:ADF}, we present the complete ADF test results on the power, cloud, and traffic datasets. The bolded results confirm the datasets' suitability for our experimental needs.

\begin{table}[tbp]
\setlength{\tabcolsep}{1.5pt}
  \centering
  \caption{ADF test results on various system load datasets. Higher Statistic and P-value indicate more pronounced dynamic characteristics.}
    \begin{tabular}{ccccc}
    \hline
    Scenario & Dataset & ADF Statistic & P-Value & Is\_Stationary \\
    \hline
    \multirow{9}[2]{*}{Power} & ETTm2 & -4.0335  & 0.0012  & T \\
          & ETTm1 & -3.8811  & 0.0022  & T \\
          & ETTh1 & -3.4880  & 0.0083  & T \\
          & ETTh2 & -3.5971  & 0.0058  & T \\
          & ECL   & -8.4449  & 0.0051  & 6 F, 315 T \\
          & \textbf{ECL\_32} & -2.4347  & 0.1322  & F \\
          & \textbf{ECL\_57} & -1.9932  & 0.2896  & F \\
          % & \textbf{ECL\_114} & -2.2084  & 0.2032  & F \\
          & \textbf{ECL\_270} & -2.8343  & 0.0535  & F \\
    \hline
    \multirow{2}[2]{*}{Cloud} & \textbf{CBW}   & -4.2933  & 0.1536  & 354 F, 476T \\
          & \textbf{ECW}   & -4.8440  & 0.0680  & 180 F, 617 T \\
          & \textbf{Switch}   & -3.3600
  & 0.0490  & 1 F, 2 T \\
          & \textbf{New APP}   & -2.3176
  & 0.2720
  & 7 F, 4 T \\
    \hline
    \multirow{4}[2]{*}{Traffic} & Traffic & -15.0210  & 0.0000  &  862 T \\ % 862
          & \textbf{Traffic\_852} & -5.0309  & 0.0000  & T \\
          & PEMS\_BAY & -26.3736  & 0.0000  &  325 T \\ % 325
          & METR\_LA & -16.0138  & 0.0000  &  207 T \\ % 207
    \hline
    \end{tabular}%
  \label{tab:ADF}%
\end{table}%

\begin{figure}[bp]
    \centering
    \includegraphics[width=\columnwidth]{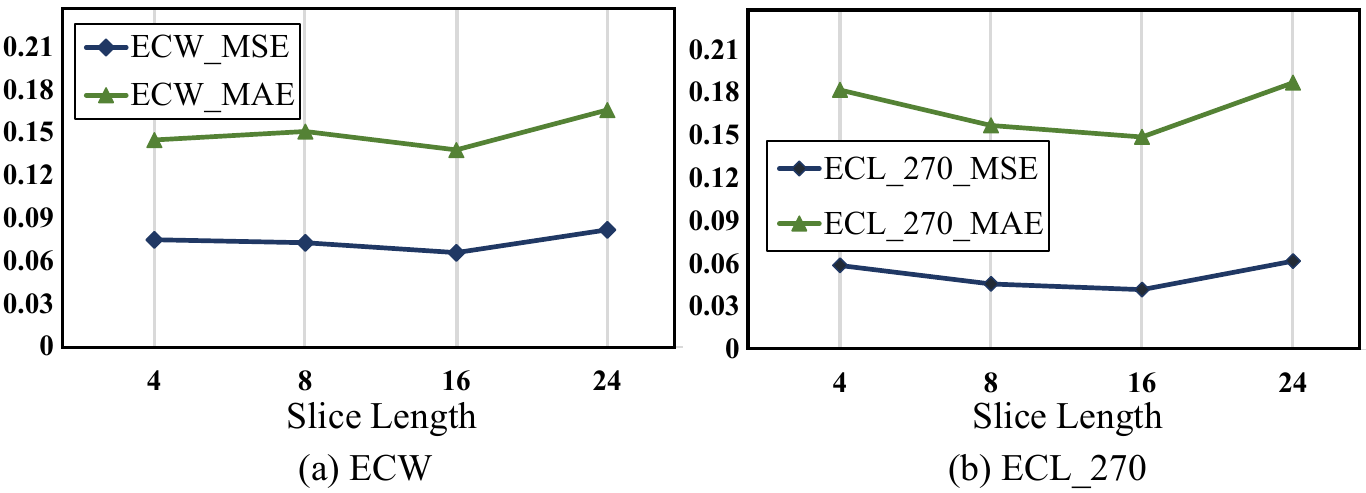}
    \caption{Prediction performance with varying slice lengths.}
    \label{fig:s_sen}
    \Description{Prediction performance with varying slice lengths.}
\end{figure}

\begin{figure*}[tbhp]
    \centering
    \setlength{\belowcaptionskip}{-0.3cm}
    \includegraphics[width=0.85\linewidth]{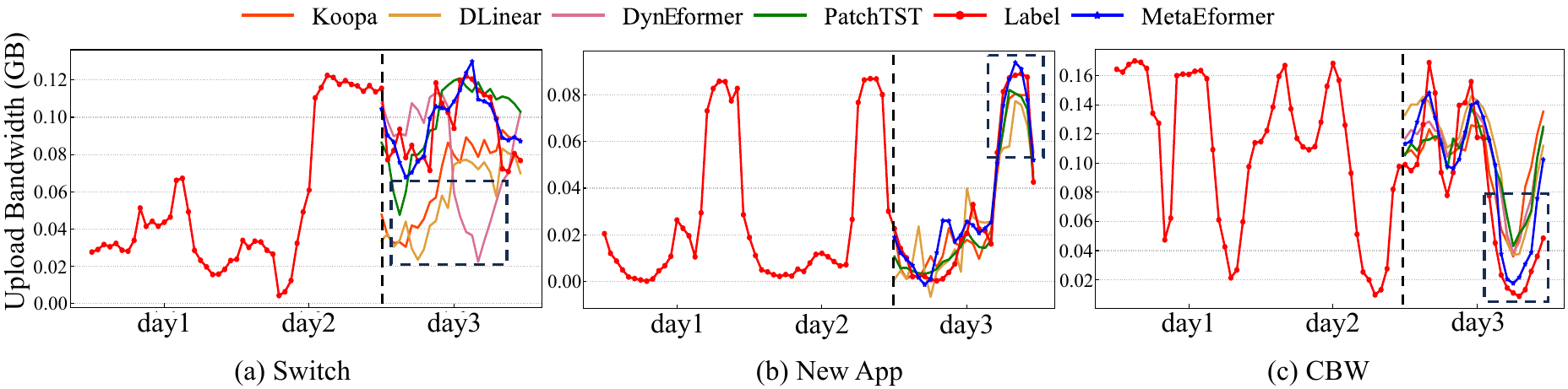}
    \caption{Prediction cases from the Switch, New App and CBW dataset.}
    \label{fig:visual}
    \Description{Prediction cases from the Switch, New App and CBW datasets.}
\end{figure*}

\begin{figure*}[htbp]
    \centering
    \includegraphics[width=0.9\textwidth]{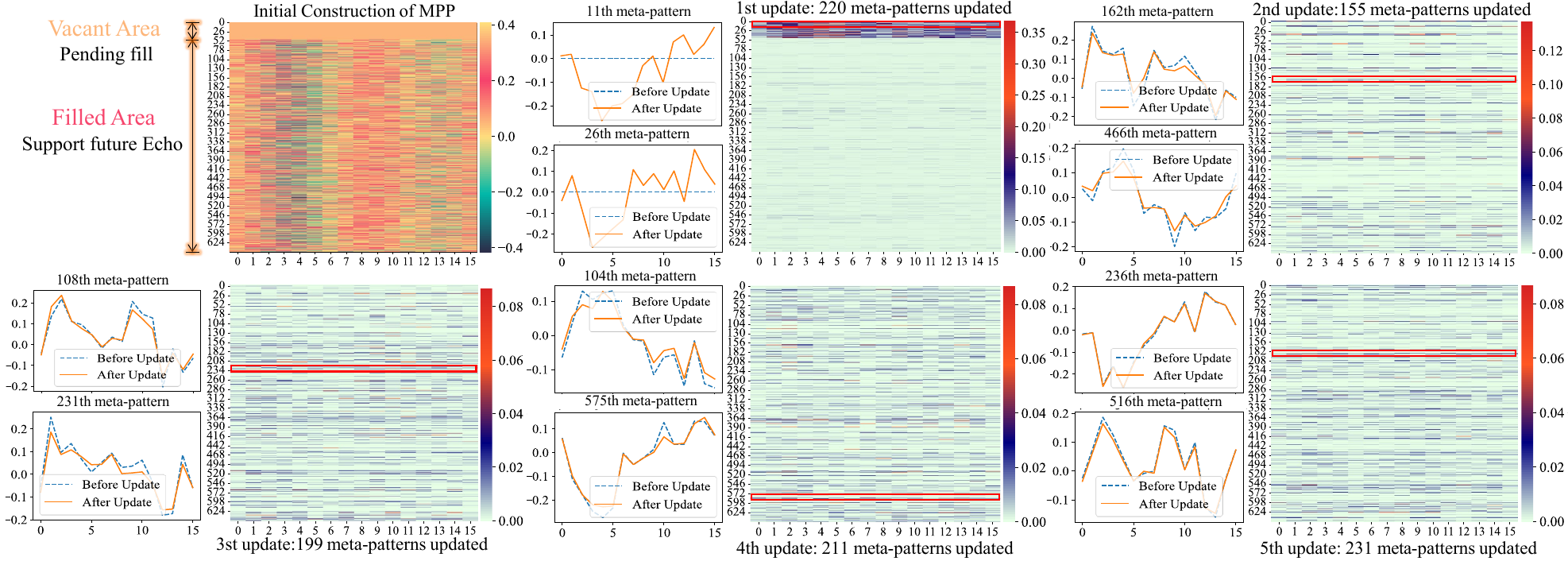}
    \caption{Visualization of MPP construction and updates. The first heatmap represents the original values of the MPP, and the subsequent heatmaps represent the difference between the updated MPP and the original. The meta-patterns with the most significant changes are visualized separately (indicated by \textcolor{red}{red} boxes).}
    \label{fig:mpp update}
    \Description{Framework overview of MetaEformer}
\end{figure*}

\begin{figure*}[htbp]
    \centering
    \includegraphics[width=0.9\textwidth]{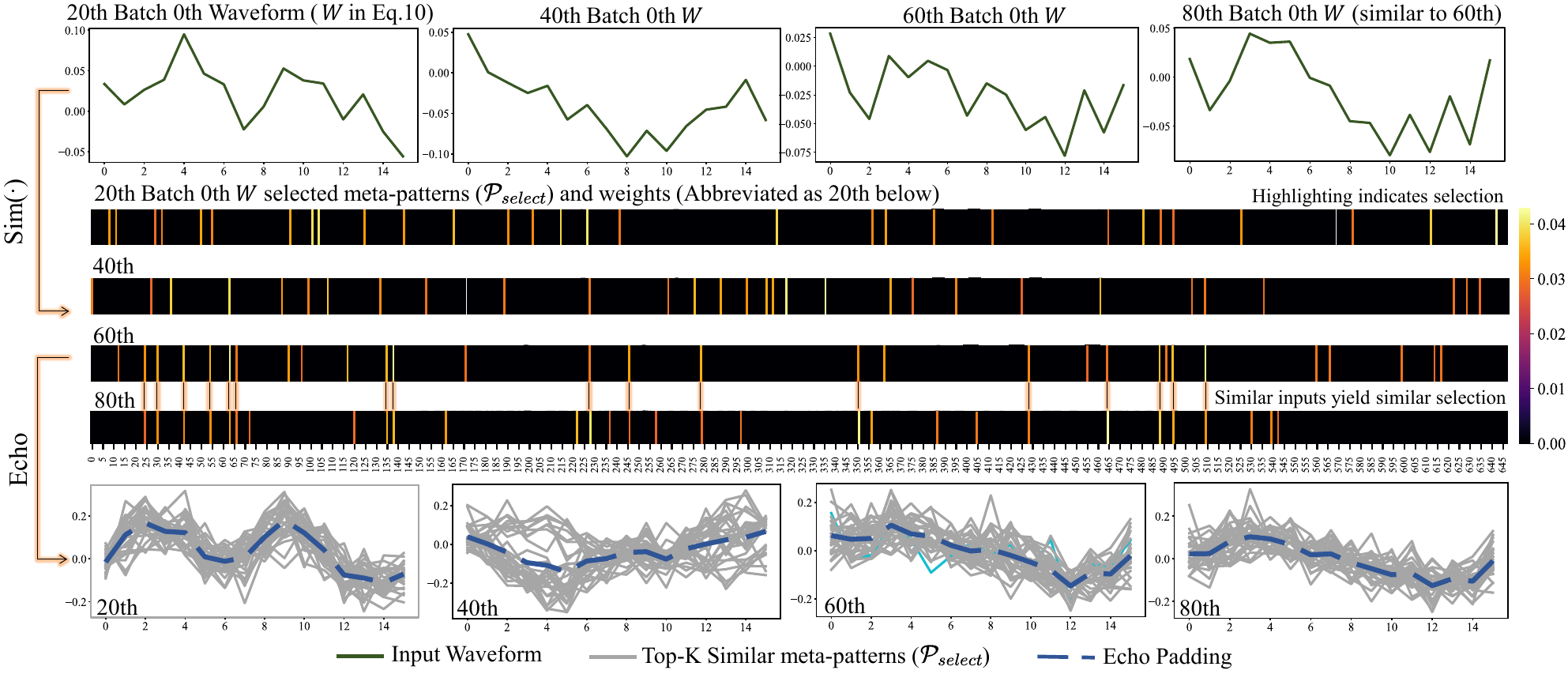}
    \caption{Displays the Echo mechanism for the input series (waveform) in every 20 batches during training on the CBW. The heatmaps depict the meta-patterns selection process within MPP, with non-selected meta-patterns set to zero. The bottom visualizes the Top-K meta-patterns $\mathcal{P}_{select}$ and the Echo Padding. Different waveforms echo meta-patterns with distinct fluctuation patterns, while similar waveforms tend to select the same meta-patterns (e.g., 60th and 80th).}
    \label{fig:Echo}
    \Description{Framework overview of MetaEformer}
\end{figure*}

\subsection{Prediction Showcases}
To assess model performance on dynamic behaviors, we compare MetaEformer with four state-of-the-art models on three dynamic datasets, as shown in Fig.~\ref{fig:visual}. The Switch and New App datasets, configured according to~\cite{10.1145/3580305.3599453}, are fully independent of the training and are used to test the models' abilities to handle concept-drift (switch behavior) and few-shot learning (new entity addition).

In the switch dataset (Fig.~\ref{fig:visual} (a)), even when high workload after the switch are provided, most models are still affected by the low workload from the previous period in the lookback window. This results in predictions much lower than the actual values. MetaEformer, using the MPP, relies more on similar meta-patterns for prediction, effectively handling switch behavior. In the new app dataset (Fig.~\ref{fig:visual} (b)), most models struggle to achieve fine-grained predictions due to the limited few-shot data, leading to inaccuracies in predicting extreme values, which are often crucial. The proposed Echo mechanism, by calculating similarity using peaks and troughs, effectively merges meta-patterns from accumulated similar cases, making it particularly adept at capturing these extreme values.

% The CBW dataset further demonstrates MetaEformer's predictive capability in more complex scenarios.

\subsection{Key Mechanisms Interpretation}
\subsubsection{Meta-pattern Pooling}\label{sec:inter MPP}
To explain the Meta-pattern Pooling mechanism, we visualize the initial construction and five subsequent updates of the MPP during MetaEformer's training on the CBW dataset using heatmaps. Each update corresponds to the full batch of data input for the current iteration. In the heatmaps, the vertical axis represents the 650 meta-patterns in the MPP (which may not be fully filled initially), and the horizontal axis shows the original series for the initial construction and the changes in each meta-pattern before and after updates. Darker areas indicate greater changes in meta-patterns. The final result is shown in Fig. \ref{fig:mpp update}.

From Fig. \ref{fig:mpp update}, it can be observed that in the early updates, due to the limited amount of data learned by the model, new meta-patterns are still being discovered and filled into the MPP during updates, as shown in the 1st updates. Significant changes occur when new patterns are introduced. The MPP quickly learns a richer and sufficient set of meta-patterns, resulting in no new meta-patterns being added. Instead, the focus shifts to updating the existing meta-patterns. This process demonstrates that the MPP update mechanism adeptly captures subtle variations in meta-patterns from subsequent input data and successfully purifies new, distinct meta-patterns.

\subsubsection{Echo Layer}\label{sec:inter Echo}
As illustrated in Fig. \ref{fig:Echo}, each Echo operation in MetaEformer actively selects suitable meta-patterns for integration based on the current input waveform, driven by the proposed similarity calculation (Sim), as highlighted in the figure. The selected meta-patterns reflect underlying structures from the original waveform while enriching it with additional global pattern information. This complementary information is crucial for the Decoder, enhancing the accuracy and robustness of the final predictions.

Throughout the MetaEformer forecasting process, the Echo Layer functions analogously to a wavelet transform. It deconstruct the original input to identify the most relevant meta-patterns (basis functions) from the MPP and determines how these meta-patterns should be integrated into the current embedding space (reconstruct). By guiding the model on both what meta-patterns are most representative and how they should be fused, the Echo Layer provides complementary information that enhances MetaEformer's ability to generate precise predictions.

\subsection{Sensitivity Analysis of Slice Length}
To investigate the impact of slice lengths $s$, we perform MetaEformer on ECW and ECL\_270 with varying $s$ = [4, 8, 16, 24]. As observed in Fig.~\ref{fig:s_sen}, the forecasting performance does not significantly vary with different values of $s$, indicating the robustness of the MetaEformer against the slice length hyperparameter.

\end{document}